\documentclass[12pt,letterpaper]{article}
\usepackage[margin=1in]{geometry}
\usepackage{amsmath}
\usepackage{graphicx}
\usepackage{enumerate}
\usepackage{amssymb}
\usepackage{url} %
\usepackage{algorithm}
\usepackage{algpseudocode}
\usepackage[doublespacing]{setspace}
\usepackage{mathdots}
\usepackage{subfig}
\usepackage{textcomp}
\usepackage{blkarray, bigstrut}
\usepackage{mathtools}
\usepackage{xcolor}
\usepackage{booktabs}
\usepackage{multirow}
\usepackage[normalem]{ulem}
\useunder{\uline}{\ul}{}
\usepackage{etoolbox}

\usepackage[numbers,sort&compress]{natbib}
\usepackage[colorlinks=true, allcolors=blue]{hyperref}

\usepackage{bm}

\usepackage{amsthm}
\newtheorem{theorem}{Theorem}[section]

\newtheorem*{lemma}{Lemma}
\newtheorem{supplemma}{Lemma}

\newcommand*{\boldone}{\text{\usefont{U}{bbold}{m}{n}1}}
\newcommand{\chLB}{\color{black}}
\title{Empirical Variational Bayes Factorization for Longitudinal Matrix Data}
\author{Jiakun Li \qquad Eric F. Lock\\[0.5em]
  \small Division of Biostatistics \& Health Data Science\\
  \small University of Minnesota Twin Cities\\
  \small Minneapolis, 55414, MN, United States}
\date{}

\begin{document}
\maketitle

\begin{abstract}
Collecting high-dimensional data over time for multiple individuals is increasingly common in biomedical research, particularly in studies of disease progression or molecular time-course experiments. Often data are missing or the timepoints are irregularly spaced across individuals. There is a lack of flexible methodology to explore latent patterns of change over time and account for missing data. We propose an empirical variational Bayes (EVB) approach to factorize high-dimensional longitudinal data, in which the underlying factors have temporal correlation over time. The EVB approach is computationally efficient and accommodates flexible covariance structures over time with automatic determination of all hyper-parameters (including the underlying rank, i.e., number of factors). Simulations demonstrate that the approach captures the underlying signal more precisely than other strategies, and performs well with respect to missing data imputation under different scenarios.  An application to a gene time-course data for multiple sclerosis treatment provides a factorization with interpretable components over time and improved imputation accuracy for missing data.
\end{abstract}


\section{Introduction}
\label{sec: intro}
Matrix factorization methods have long been critical to analyze and summarize high-dimensional data for diverse tasks such as latent structure inference, data compression, and missing data imputation. %
Recently, empirical variational Bayes has been shown to be a powerful, flexible, and efficient approach to low-rank matrix factorization. Under a variational Bayes approach the true posterior for a Bayesian model is approximated by a variational distribution with simplified assumptions \citep{fox2012tutorial}, under EVB the model hyperparameters are also empirically determined.  Previous work describes a simple EVB approach to matrix factorization that results in an efficient closed-form low-rank solution \cite{nakajima13a}.
Empirical Bayes matrix factorization flexibly incorporates different forms of priors, allowing novel combinations of classical models such as sparse matrix factorization and non-negative matrix factorization \cite{EBMF}.
In the context of multiple linked matrices, EVB has been used to decompose signal into low-rank approximations for shared and unshared structure \cite{EV_BIDI}.

Advancements in biotechnology have enabled high-throughput data collection (e.g., for ``omics" or imaging modalities) from multiple timepoints,  motivating longitudinal studies. However, in most cases it is difficult to obtain complete data, often caused by patient unavailability or dropout. This can lead to several issues, such as biased results and a reduction in statistical power, compromising the validity of the study and the reliability of its findings. As a motivating example, consider a study that investigates genomic responses of multiple sclerosis (MS) patients to the genetic therapy recombinant human interferon beta (rIFN$\beta$) \citep{IFNB}. The study involves analyzing a gene time course dataset that measures 76 gene expression levels in $53$ subjects, at $7$ timepoints over a 2-year followup period. However, $10.5\%$ of the data are missing across all timepoints, limiting the validity of inference.

Several classes of methods have been developed to apply factorization models to longitudinal data settings. One popular area of research are tensor factorization methods \citep{kolda2009tensor}. The tensor framework generalizes the two-way matrix to higher-order arrays.  Tensor factorization methods enjoy many desirable properties of matrix factorization, such as low rank representation of the data and interpretable additive components. They can conveniently accommodate longitudinally collected high-dimensional matrices by adding time as a third dimension, forming a third-order tensor array. Tensor factorization methods have been developed specifically to accommodate temporal structure, by incorporating functional representations or smoothing along the temporal dimension \citep{guan2024smooth, TEMPTED}. Other tensor methods have been developed for imputation in a longitudinal setting using a tensor factorization \citep{BAMITA}.  However, treating time as an additional dimension within a traditional tensor framework assumes that timepoints are aligned across individuals. %

{\chLB  For matrices, a related class of methods incorporate auxiliary side information on rows or columns to inform matrix factorization. For example, \cite{singh2008relational} considers additional matrices of covariates for the rows and columns, and borrows information to factorize all simultaneously.  Variational Bayes has enabled computationally efficient posterior approximation for matrix factorization under structured Bayesian priors that incorporate side information for rows or columns \citep{gonen2014kernelized}; however, this requires model-specific hyperparameters that are not empirically determined. More recently, \cite{MFAI} describe an empirical variational Bayes approach in which auxiliary covariates inform the latent factor means.  While time can be incorporated as a covariate in such methods to capture mean trends over time, they do not inherently accommodate within-individual correlation structures over time that are common for longitudinal studies. }  

{\chLB A small but growing number} of matrix factorization models {\chLB are explicitly designed} for longitudinal settings. For example, \cite{latent_factor_lmm} use a factor analysis approach and models the scores using a multivariate linear mixed model. This induces correlation within individual observations, but limits flexibility to model the correlation to the specific form under a linear mixed model. Bayesian methods, on the other hand, can deal with more complex correlation structures by naturally incorporating flexible priors for the factors. For example, \cite{bayes_factor_lu_etal} uses a factor model for random effects within a mixed modeling framework, where the scores are employed with an infinite mixture prior. However, this work centered on the task of clustering instead of signal estimation or imputation, which is the focus of this paper.  Alternatively,  \cite{wang2024multilayer} propose a longitudinal factor model for disease progression from high-dimensional data, but assume a transition between two latent states over time rather than a more general correlation structure.

We propose an empirical variation Bayes factor model for high-dimensional longitudinal data that simultaneously addresses estimation and imputation, even when timepoints are inconsistently measured. Our contribution has four main components. 
First, we introduce a general repeated measures model for high-dimensional data using latent factors that flexibly accommodates a range of correlation structures over time. 
Second, we use an EVB framework for estimation that does not require *any* user-specified tuning parameters, and that achieves automatic rank selection—an important advantage since manual rank selection can be difficult or computationally prohibitive for large datasets. Third, we develop a novel empirical Bayes random process model (EBRPM), which extends prior work to allow efficient computation and unified estimation of both priors and variationally approximated posteriors. This formulation leverages empirical Bayes principles to maximize efficiency while retaining the ability to recover complex signals in high-dimensional settings. Fourth, our framework naturally handles missing data, including both entrywise and blockwise missingness, by retaining a key Bayesian property: the sharing of information across individuals to estimate a common prior distribution. This feature is especially important for blockwise missingness, where data for an individual may be entirely absent at a timepoint and imputation without borrowing information and temporal correlation is impossible.

\section{Methods} 
\label{Sec: Methods}

\subsection{Model Construction}
\label{sec:model}

Here we describe our model for aligned measurement times across all subjects; extensions to scenarios with irregularly measured timepoints across subjects are straightforward, and are described in Section~\ref{Sec: Irreg Model} of the Supplementary Materials.  %

Assume there are $n$ subjects, each with $p$ variables measured at $t_{1}, \dots, t_{D}$ timepoints. We define the rank $R$ model as
\begin{equation} \label{equn: model def}
    \bm Y_t = \bm U \bm V_t^T + \bm E_t, \: t = t_1,\dots,t_D
\end{equation}
where $\bm Y_t \in \mathbb{R}^{p \times n}$ are the observed data at each timepoint, $\bm U \in \mathbb{R}^{p \times R}$, $\bm V_t \in \mathbb{R}^{n \times R}$ are factor matrices and $\bm E_t$ is an error matrix at time $t$. %
Following standard terminology, we refer to $\bm U$ as the \textit{loading} matrix, $\bm V$ as the \textit{score} matrix, and $R$ as the rank/number of factors. The model can be expanded as the sum $R$ rank-1 components, via the outer products between columns of $\bm U$ and $\bm V_t$,
\begin{equation} \label{component rep model}
    \bm Y_t = \sum_{r=1}^R \bm{u}_r \bm{v}_{tr}^T + \bm E_t
\end{equation}
where $\bm{u}_r$ and $\bm{v}_{tr}$ denote the $r$-th column of $\bm U$ and $\bm V_t$, respectively. %
Following a Bayesian framework, we give independent normal priors to each column of $\bm U$, and a joint normal prior to $\tilde{\bm{v}}_{ri} \coloneq [v_{t_1ri}, \dots, v_{t_Dri}]$, the corresponding entries of $\bm V_t$ across timepoints:
\begin{equation} \label{equn: prior}
    \bm{u}_{rj} \overset{iid}{\sim} N(0,\eta_r^2), \;
    \tilde{\bm{v}}_{ri} \overset{iid}{\sim} mvN(0,\bm K_r), \;
    E_{tji} \overset{iid}{\sim} N(0,\sigma_{tji}^2)
\end{equation}
where $\bm K_r$ captures component-specific temporal correlation in the factor scores, as detailed in Section~\ref{sec: covariance}. {\chLB The use of normal priors with factor-specific variances can be considered an automatic relevance determination (ARD) prior \citep{Bishop_BPCA, Variational_PC, zhao2015bayesian, nakajima2015condition}.}

\subsection{Empirical Variational Bayes Estimation}
We begin by describing the estimation algorithm for the rank $1$ model, and generalize to the multiple rank scenario in Section \ref{sec:rank_r}. In a standard empirical Bayes framework, model estimation proceeds by first obtaining point estimates for the prior parameters $\eta^2$, $\bm K$, $\bm{\sigma}^2$, e.g., by maximizing the marginal log-likelihood under estimated priors $g(\cdot)$:
\begin{equation} \label{equn: marg lkhd}
    \begin{aligned}
        (\hat{\eta},\hat{\bm K}, \hat{\bm{\sigma}}^2) %
        &= \underset{\bm \eta,\bm K,\bm{\sigma}^2}{\arg \max} \: \log \int \int \prod_{t=t_1}^{t_D} p(\bm Y_t| \bm{u}, \bm{v}_t, \bm \sigma^2)g(\bm{v}_t|\bm K) g(\bm{u}|\bm \eta) d\bm{v}d\bm{u},
    \end{aligned}
\end{equation}
then performing inference on $\bm{u}$, $\bm{v}$ via the posterior distribution
\begin{equation} \label{equn:full_pos_dist}
    p(\bm{u},\bm{v}|\bm Y,\hat{\eta},\hat{\bm K},\hat{\bm{\sigma}}^2).
\end{equation}

\subsubsection{Approximate Inference}
The integrals involved in both \eqref{equn: marg lkhd} and \eqref{equn:full_pos_dist} are intractable %
due to marginalizing over latent variables. Therefore, we must resort to approximate inference techniques. Markov Chain Monte Carlo (MCMC) \citep{MCMC} is the most common technique where approximate samples are drawn from the true posterior, however it often suffers from high computational demands for large-scale problems. Instead, we propose a computationally attractive variational Bayes solution for our setting. %

In variational Bayes, posterior estimation is posed as an optimization problem, where we search for a factor distribution $q(\bm u,\bm{v})$ that gives the smallest KL divergence with the true posterior: \\ $q^*(\bm u,\bm{v}) = \underset{q \in \mathcal{Q}}{\arg\min} \: \text{KL}(q(\bm u,\bm{v})|p(\bm u,\bm{v}|\bm Y))$ \citep{VI_Review}. This is done by maximizing the evidence lower bound (ELBO),
\begin{equation}
    \begin{aligned}
        F_{ELBO}(q, \eta, \bm K, \bm{\sigma}) &\coloneq E_{q}[\log \frac{p(\bm Y,\bm u,\bm v; \eta, \bm K,\bm{\sigma})}{q(\bm u,\bm v)}] \\
        & = l(\eta,\bm K,\bm{\sigma}) - \text{KL}(q_{u,v}||p_{u,v|\bm Y,\eta,\bm K,\bm{\sigma}})
    \end{aligned}
\end{equation}
which is a lower bound for the log marginal likelihood, as the KL divergence is non-negative. Since $l(\eta,\bm K,\bm{\sigma})$ is constant with respect to $q(\bm{u},\bm{v})$, $\underset{q}{\arg\max} \: F(q, \eta, \bm K, \bm{\sigma}) = \underset{q}{\arg\min} \: \text{KL}(q_{\bm u,\bm{v}}||p_{u,\bm{v}|Y,\eta,\bm K,\bm{\sigma}})$. However, without further constraints on $q$, this optimization is difficult. We restrict the space of factor distributions to the mean field variational family,
\begin{equation}
    \begin{aligned}
        \mathcal{Q} = \{q(\bm u,\bm{v}) : q(\bm u,\bm{v}) = \prod_{j=1}^p q_{u_j}(u_j) {\chLB \prod_{i=1}^n q_{\tilde{\bm{v}}_i}(\tilde{\bm{v}}_i)} \}
    \end{aligned}
\end{equation}
Under this factorization assumption, we develop a coordinate ascent algorithm to solve for $q^*(\bm u,\bm{v})$ given by Algorithm \ref{alg:rank_1}. To allow for modeling of flexible correlation structure, we define a simpler, more general empirical Bayes model with optimization in the following section that can be used to solve step \ref{alg:1.4} and step \ref{alg:1.5} of Algorithm \ref{alg:rank_1}.
\begin{algorithm}
\caption{Coordinate Ascent Model Estimation with Rank-$1$ Model}\label{alg:rank_1}
\begin{algorithmic}[1]
    \Require{Initialize $q_u(\bm{u})^{(0)}$, $q_v(\bm{v})^{(0)}$, $\eta^{(0)}$, $\bm K^{(0)}$, $\bm{\sigma}^{(0)}$}
    \Repeat
        \State $m \gets m + 1$
        \State $\bm{\sigma}^{(m)} \gets \underset{\bm{\sigma}}{\arg \max} \: F_{ELBO}(q_{\bm{u}}^{(m-1)}, q_{\bm{v}}^{(m-1)}, \eta^{(m-1)}, \bm K^{(m-1)}, \bm{\sigma})$
        \State $q_{\bm u}^{(m)}, \eta^{(m)} \gets \underset{q_{\bm u}, \eta}{\arg \max} \: F_{ELBO}(q_{\bm{u}}, q_{\bm{v}}^{(m-1)}, \eta, \bm K^{(m-1)}, \bm{\sigma}^{(m)})$ \label{alg:1.4}
        \State $q_{\bm v}^{(m)}, \bm K^{(m)} \gets \underset{q_{\bm v}, \bm K}{\arg \max} \: F_{ELBO}(q_{\bm{u}}^{(m)}, q_{\bm{v}}, \eta^{(m)}, \bm K, \bm{\sigma}^{(m)})$ \label{alg:1.5}
    \Until{convergence criterion met}
    \State \textbf{Return} $q^{(m)}_{\bm u}, q^{(m)}_{\bm v}, \eta^{(m)}, \bm K^{(m)}, \bm{\sigma}^{(m)}$
\end{algorithmic}
\end{algorithm}

\subsubsection{Optimizing via Empirical Bayes Random Process Means (EBRPM) Model} \label{sec:EBRPM}

In this section, we define the Empirical Bayes Random Process Means Model (EBRPM). This model is closely related to the proposed model, in the sense that there is an equivalence with respect to $\bm u$ or $\bm v$ when the other is fixed. Suppose that we have noisy observations $\bm{x}_1, \dots, \bm{x}_n$ of true values $\bm{\theta}_1, \dots, \bm{\theta}_n$, each $\bm{x}_i \in \mathbb{R}^{d_i}$ can potentially be of different dimension: 
\begin{equation} \label{equation: EBRPM_1}
   \bm{x}_i | \bm{\theta}_i \overset{ind}{\sim} \mathcal{N}_{d_i}(\bm{\theta}_i, \bm S_i)
\end{equation}
The $\bm S_i$'s are known diagonal matrices reflecting independent noise and $\bm{\theta}_i$ are generated from a random process. Here we focus on modeling correlation within the $\bm{x}_i$'s and restrict our prior family for $\bm{\theta}_i$ to zero-mean Gaussian processes,
\begin{equation} \label{equation: EBRPM_2}
   \bm{\theta}_i \overset{iid}{\sim}\mathcal{GP}(0,\bm K(\cdot,\cdot))
\end{equation}
since there are many well studied covariance functions and this distribution generally have good computational properties. We are interested in estimating the $\bm{\theta}_i$'s and $\bm K$ via an empirical Bayes approach, and this is performed in two steps:

1. Estimate $\bm K$ by type II maximum likelihood:
\begin{equation} \label{equn: max_K}
    \hat{\bm K} = \underset{\bm K}{\arg \max} \int \prod_{i=1}^n p(\bm{x}_i | \bm{\theta}_i)g(\bm{\theta}_i|\bm K) d\bm{\theta} = \underset{\bm K}{\arg \max} \prod_{i=1}^n \mathcal{N}_{d_i}(\bm{x}_i | 0,\bm S_i + \bm K)
\end{equation}

2. Calculate the posteriors for $\bm{\theta_i}$'s using the estimated prior $\hat{\bm K}$ from the first step:
\begin{equation}
p(\bm{\theta}_i|x,\hat{g}) \propto p(\bm{x}_i| \bm{\theta}_i) g(\bm{\theta}_i|\hat{\bm K})
\end{equation}

This model is a special case of the empirical Bayes multivariate normal means model (EBMNM) \citep{EBMNM} with multivariate Normal priors when $d_i = d_j$ for all $i,j$ realized at the same timepoints, and the empirical Bayes normal means model (EBNM) \citep{EBMF} with normal priors when $d_i=1$ for all $i$. 
Following the EBNM notation, we summarize the estimation procedure as a function, with an input of $(\bm{X},\underset{\sim}{\bm S})$ and output $(p_{\bm{\theta}|\bm{x}},\hat{\bm K})$, where $\bm X = [\bm x_1,\dots,\bm x_n]$, $\underset{\sim}{\bm S} = [\bm S_1,\dots,\bm S_n]$. In Theorem~\ref{thrm EB VB equiv} we establish that steps $4$ and $5$ of Algorithm \ref{alg:rank_1} can be solved by an EBRPM problem.
\begin{theorem} \label{thrm EB VB equiv}
    Step 4 and 5 of Algorithm \ref{alg:rank_1} are equivalent to the following EBRPM problems:
    \begin{equation} \label{alg:1.4 EB}
        \underset{q_{\bm u}, \eta}{\arg \max} \: F_{ELBO}(q_{\bm{u}}, q_{\bm{v}}, \eta, \bm K, \bm{\sigma}) = \text{EBRPM}(\bm{x}_j = \frac{\sum_{t,i} Y_{tji} \overline{v}_{ti}/\sigma^2_t}{\sum_{t,i} \overline{v_{ti}^2}/\sigma^2_t}, \bm S_j = \frac{1}{\sum_{t,i} \overline{v_{ti}^2}/\sigma^2_t})
    \end{equation}
    \textit{and}
    \begin{equation} \label{alg:1.5 EB}
        \underset{q_{\bm v}, \bm K}{\arg \max} \: F_{ELBO}(q_{\bm{u}}, q_{\bm{v}}, \eta, \bm K, \bm{\sigma}) = \text{EBRPM}(\bm{x_i} = \frac{\sum_{j=1}^p \overline{u}_j \bm{y}_{ji}}{\sum_{j=1}^p \overline{u_j^2}}, \bm S_i = \frac{1}{\sum_{j=1}^p \overline{u_j^2}} \bm{\Sigma})
    \end{equation}
\end{theorem} 
\noindent
\textit{Where} $\bm{y}_{ji} = [Y_{t_1 ji}, \dots, Y_{t_D ji}]^T$, $\bm{\Sigma} = \text{diag}(\sigma_{t_1}^2, \dots, \sigma_{t_D}^2)$, $\overline{u}, \overline{v}$ \textit{represent the first and second moments of} $\bm{u}, \bm{v}$ ]\textit{under} $q_{\bm u}, q_{\bm v}$.

 The proof of Theorem~\ref{thrm EB VB equiv} is provided in Section~\ref{sec:proof} of the Supplementary Materials, and additional computational details are provided in Sections~\ref{app:ebrpm} and \ref{app:elbo}. 
 
\subsection{Extension to Rank \textit{R} Model}\label{sec:rank_r}

For the rank $R$ model ($R>1$), we simply extend the variational approximation by setting factor distributions for all $\bm{u}_r$, {\chLB $\tilde{\bm{v}}_{ri}$} as in (\ref{component rep model}). $F_{ELBO}$ is then maximized over $(q_r, \eta_r, \bm K_r)_{r=1}^R, \bm{\sigma}$ and Algorithm \ref{alg:rank_1} is modified accordingly to iteratively update each factor component while keeping all others constant. The update steps for each $\bm{u}_r$, $\tilde{\bm{v}}_{ri}$, $r=1,\dots,R$ follows Theorem \ref{thrm EB VB equiv}, with $\bm Y_t$ replaced by $\bm Y_t - \sum_{r' \neq r} \bm{u}_{r'} \bm{v}_{tr'}$. We summarize the estimation prodecure in Algorithm \ref{alg:rank_R}.

\begin{algorithm}
\caption{Extension to Rank $R$ Model}\label{alg:rank_R}
\begin{algorithmic}[1]
   \Require{Initialize $q_{u_r}(\bm{u}_r)^{(0)}$, $q_{v_r}(\bm{v}_r)^{(0)}$, $\eta_r^{(0)}$, $\bm K_r^{(0)}$, $\bm{\sigma}^{(0)}$}, $r=1,\dots,R$
   \Repeat
       \State $m \gets m + 1$
       \State $\bm{\sigma}^{(m)} \gets \underset{\bm{\sigma}}{\arg \max} \: F(q_{\bm{u}}^{(m-1)}, q_{\bm{v}}^{(m-1)}, \eta^{(m-1)}, \mathbb{K}^{(m-1)}, \bm{\sigma})$
       \For{$r=1,\dots,R$}
           \State $q_{\bm u_r}^{(m)}, \eta_r^{(m)} \gets \underset{q_{\bm u_r}, \eta_r}{\arg \max} \: F(q_{\bm{u}}, q_{\bm{v}}^{(m-1)}, \bm{\eta}, \mathbb{K}^{(m-1)}, \bm{\sigma}^{(m)})$ \label{alg:2.4}
           \State $q_{\bm v_r}^{(m)}, \bm K_r^{(m)} \gets \underset{q_{\bm v_r}, \bm K_r}{\arg \max} \: F(q_{\bm{u}}^{(m)}, q_{\bm{v}}, \bm{\eta}^{(m)}, \mathbb{K}, \bm{\sigma}^{(m)})$ \label{alg:2.5}
       \EndFor
   \Until{convergence criterion met}
   \State \textbf{Return} $q_{\bm {u}_r}, q_{\bm {v}_r}, \eta_r, \bm K_r, \bm{\sigma}, r=1,\dots,R$
\end{algorithmic}
\end{algorithm}

When $R$ is large, this algorithm can become computationally intensive, as all $R$ factors are updated at each iteration. To alleviate this issue, we implement %
a greedy algorithm, where rank-$1$ models are sequentially fitted to the residuals from the sum of previous estimated models, effectively adding one column to the factor matrices at a time, as described in Algorithm~\ref{alg:greedy}. 

\begin{algorithm}
\caption{Greedy Algorithm for Initialization}\label{alg:greedy}
\begin{algorithmic}[1]
   \Require{Initialize $R=1$ and $\bm S_t = 0$ for $t=t_1,\dots,t_D$}
   \Repeat
       \State Set $\bm R_t = \bm Y_t - \bm S_t$ for $t=t_1,\dots,t_D$
       \State Perform Algorithm \ref{alg:rank_1} with $\bm Y_t = \bm R_t$ for $t=t_1,\dots,t_D$
       \State Set $\hat{\bm{u}}_R = E_{q_u}[\bm U]$, $\hat{\bm{v}}_{tR} = E_{q_v}[\bm V_t]$ for $t=t_1,\dots,t_D$
       \State Set $\bm U = [\bm{u}_1 \cdots \bm{u}_R]$, $\bm V_t = [\bm{v}_{t1} \cdots \bm{v}_{tR}]$, $\bm S_t = \bm U \bm V_t^T$ for $t=t_1,\dots,t_D$
       \State $R = R+1$
    \Until{convergence criterion met}
    \State \textbf{Return} $\bm U, \bm V_t, \bm S_t$ for $t=t_1,\dots,t_D$
\end{algorithmic}
\end{algorithm}

In practice, Algorithm \ref{alg:rank_R} often achieves significantly better fit, given that it optimizes $F_{ELBO}$ over a larger space. 
Combining both by using the greedy algorithm as an initialization for the coordinate ascent algorithm results in significantly less iterations required to achieve convergence, thus reducing computational cost without sacrificing model fit.

\subsection{Covariance Functions}
\label{sec: covariance}

Several covariance functions for the Gaussian process prior of $\bm K$ have been developed for a wide range of scenarios. We implement several common covariance functions, %
as follows: 
\begin{align}
  \textit{White noise kernel (Indep)}: \; \;  \bm K(s,t) &= \sigma_v^2 \boldone(s=t) \label{wn:covariance} \\
  \textit{Exchangable covariance (EC)}: \; \; \bm K(s,t) &= \rho^{\boldone(s\neq t)} \sigma_v^2 \label{cov.ec}\\
  \textit{Radial Basis Function kernel (RBF)}: \; \; \bm K(s,t) &= \sigma_v^2 \exp(-\frac{\Vert s-t \Vert^2}{2\alpha^2}) \label{cov.rbf}\\
    \textit{Free covariance (Free)}: \; \;  \bm K(s,t) &= \sigma_{st}^2, \: \bm K \in \mathcal{U+} \label{free:covariance}
\end{align}
where $\mathcal{U}+$ denotes the set of positive definite matrices. We consider the model with independence over time as a benchmark to quantify the performance gain of modeling covariance. This is a special case of the model described in \cite{nakajima13a} and \cite{EV_BIDI}, which has a closed form analytic solution to the empirical variational Bayes approximation. Thus we opt to use their code for faster implementation.  Covariance \eqref{free:covariance} is a free structure with a different covariance for each pair of timepoints. It is impractical at highly irregular timepoints, but is the most flexible structure for the regular timepoint model, and is likely to perform well when the true model structure is unclear.  The white noise and exchangeable covariances \eqref{wn:covariance} and \eqref{cov.ec} have a closed form solution for their parameters in the optimization sub-step \eqref{equn: max_K}, while the other covariance structures may be estimated via gradient-based techniques (see Section~\ref{app:ebrpm} of the Supplementary Materials for more details). 

{\chLB The choice of covariance structure may depend on context and nature of the data. For example} the RBF covariance \eqref{cov.rbf} accounts for distance between timepoints and is well-motivated when observations are irregularly spaced over time.  {\chLB Otherwise, our results show that the \textit{free} covariance model is generally a robust choice to misspecification of the true covariance structure (see simulations in Section~\ref{Sec: Simulation} and Section~\ref{sec:misspec}), and is a flexible option.  Alternatively, cross-validation of imputation accuracy may be used to select a covariance structure, as done for our application in Section~\ref{Sec: Real Data}. The objective function $F_{ELBO}$ itself is not recommended as a general criteria to select covariance, as the empirical estimation will always favor the \textit{free} model because it subsumes the other models. }

\subsection{Model Dimension Selection}
\label{sec:rank_selection}
Variational Bayes matrix factorization methods have the advantage of automatically selecting the rank concurrently with the estimation framework \citep{EBMF,Variational_PC}. This comes from the fact that $F_{ELBO}$ can be viewed as a balance between model fit and prior resemblence \citep{VI_Review}:
\begin{equation}
    \begin{aligned}
        F_{ELBO} %
        &\propto \sum_{t=t_1}^{t_D} -\frac{1}{2\sigma_t^2} E_{q}[\|\bm Y_{t} - \bm U \bm V_t^T\|_F^2] - \text{KL}(q(\bm U,\bm{V}) || g(\bm U,\bm{V}))
    \end{aligned}
\end{equation}
Let $\bm U \in \mathbb{R}^{p\times R_1},\bm{V}_t \in \mathbb{R}^{n\times R_1}$ and a larger model $\bm U' \in \mathbb{R}^{p\times R_2},\bm{V}'_t \in \mathbb{R}^{n\times R_2}$. It is clear that when $R_2>R_1$, $\text{KL}(q_{\bm u', \bm v'} || g_{\bm u', \bm v'}) > \text{KL}(q_{\bm u, \bm v} || g_{\bm u, \bm v})$. So when the decrease in expected squared residuals is less than the penalty incurred by the KL divergence between the added factor distributions and their priors, $F_{ELBO}$ cannot increase and prefers the lower rank model, %
equivalent to optimizing $q(\bm U,\bm{V})$ towards point masses at zero. Thus, setting $R$ %
large initially, the backfit algorithm ``prunes" the model by setting redundant factors to zero.

Further, rank sparsity is also achieved through the empirical Bayes step, which can minimize the prior density $g(\cdot)$ to a point mass at zero. As noted in \cite{EBMF} and \cite{Variational_PC}, it is possible that the marginal likelihood is maximized for $g(\cdot)$ at zero, and the following lemma from \cite{EBMF}  shows that $F_{ELBO}$ is guaranteed to be maximized at $g(\cdot)$ if the marginal likelihood is, which can be applied to our model as well.

\begin{lemma}
    For a rank $1$ model, if \[
\underset{g_u,g_v}{\arg\max} \: l(q_{\bm u},q_{\bm v},g_{\bm u}, g_{\bm v},\bm{\sigma}) = (\delta_{g_u}, \delta_{g_v}),
\]
    then \[
\underset{g_u,g_v}{\arg\max} \: F(q_{\bm u},q_{\bm v},g_{\bm u}, g_{\bm v},\bm{\sigma}) = (\delta_{g_u}, \delta_{g_v})
\] 
\end{lemma}

In our context, this means that $\bm{\eta},\mathbb{K}$ are maximized at $0$. This justifies the procedure of the greedy algorithm, as columns will be added to the factor matrices until eventually adding a column of zeros which terminates the algorithm.

\subsection{Identifiability}
\label{sec:identifiability}
{\chLB As in other matrix factorization models, the loading and score components are only identifiable up to a scaling factor, and permutations \citep{EBMF,Hoyer2004}.
Specifically, for any $c_r \neq 0$, rescaling $\bm u_r \mapsto c_r\bm u_r$ and $\tilde{\bm v}_{ri} \mapsto \tilde{\bm v}_{ri}/c_r$, together with $\eta_r \mapsto |c_r|\eta_r$, $\bm K_r \mapsto \bm K_r/c_r^2$, leaves both $F_{ELBO}$ and the estimated signal $E_q[\bm U\bm V_t^T]$ invariant. This is generally not an issue as signal estimation and imputation are unaffected. A scale convention can be imposed by normalizing each loading estimate to norm 1 and adjust the scale of the corresponding scores accordingly.}

\subsection{Missing Data Imputation}
\label{Sec: Imputation}
Here we consider incomplete data and the task of imputing missing elements. Denote the indices of missing entries as $\mathcal{M} = \{ (t,j,i): Y_{tji} \text{ not observed} \}$ and $Y_{\mathcal{M}} = \{Y_{tji}:(t,j,i) \in \mathcal{M}\}$, we propose an Expectation-Maximization type approach, summarized in Algorithm \ref{alg:impute}.

\begin{algorithm}
\caption{Imputation}\label{alg:impute}
\begin{algorithmic}[1]
    \State Set $Y^{(0)}_{tji} = Y_{tji} \cdot \boldone((t,j,i) \notin \mathcal{M})$
    \Repeat
        \State $k \gets k + 1$
         \State Perform the complete data algorithm on $\bm Y_{t}^{(k-1)}$ and denote $\hat{\bm S}_t^{(k)} = E_{q}[\bm U \bm V_t^T]$ for $t=t_1,\dots,t_D$
         \State Impute $\bm Y^{(k)}_{tji} = \hat{\bm S}^{(k)}_{tji}$, $(t,j,i) \in \mathcal{M}$
    \Until{$\sum_{(t,j,i) \in \mathcal{M}} (\hat{\bm S}^{(k)}_{tji} - \hat{\bm S}^{(k-1)}_{tji})^2 < \epsilon$} for some small $\epsilon$
    \State \textbf{Return} $\bm Y^{(k)}_{t_1},\dots, \bm Y^{(k)}_{t_D}$
\end{algorithmic}
\end{algorithm}

This approach is analogous to previous methodologies including \cite{EV_BIDI} and \cite{softimpute}. Note that step 4 and 5 are respectively the ``maximization" and ``expectation" steps, as the former maximizes $F_{ELBO}$ according to Theorem \ref{thrm EB VB equiv} and Algorithm \ref{alg:rank_R}, and the latter replaces the missing entries with the expected value under current estimates of the factor distributions. This algorithm is justified by Theorem \ref{thrm: imputation}, which shows the imputation algorithm optimizes $F_{EBLO}$ over missing entries $\bm Y_{\mathcal{M}}$ as well as the latent factor distributions and hyperparameters. 
\begin{theorem} \label{thrm: imputation}
    Algorithm \ref{alg:impute} iteratively maximizes $F_{ELBO}$ over $q(\bm U,\bm V),$ $\bm{\sigma}, \bm K, \bm{\eta}, \bm Y_\mathcal{M}$
\end{theorem}
The proof of Theorem \ref{thrm: imputation} is given in Section~\ref{sec:proof} of the Supplementary Materials. Compared to the related imputation algorithm in \cite{EV_BIDI}, we do not require pre-normalizing the data by estimating $\bm{\sigma}$ first, but rather incorporate it into the optimization algorithm. This also avoids handling $\bm \sigma$ differently under entrywise and blockwise missingness.

\section{Simulations} \label{Sec: Simulation}

We conduct two simulation studies to assess both signal recovery and imputation accuracy. We use the regular timepoints model in all simulations to allow for comparisons with competing methods that may not accommodate irregular timepoints. 

\subsection{Simulation study 1: Complete data signal recovery}
\label{sec:complete_sim}
Here we assess our model's ability to estimate underlying structure for complete data, and illustrate aspects of the different covariance functions. We generate matrices of size $100 \times 120$ for 5 regularly spaced timepoints with true rank $3$ according to our model description, under each of \textit{EC}, \textit{RBF}, and \textit{Free} covariance structures with varying signal-to-noise ratios (SNR) and correlation levels. We also use a block covariance structure that none of the covariance functions correctly capture to assess misspecification, where an $EC$ correlation is given independently for different subsets of timepoints. See Section~\ref{app:sims} of the Supplementary Materials for complete details on the data generation for each scenario. We implement our model using the three non-trivial covariance functions in Section~\ref{sec: covariance} (EC, RBF, Free), and without correlation using the EVB implementation in \cite{EV_BIDI}. We also consider the general tensor decomposition method TEMPTED that doesn't explicitly model covariance between timepoints but rather treats time as a continuous variable \citep{TEMPTED}. Accuracy is measured using the relative squared error:
\begin{equation}
    \text{RSE} = \sum_{t,j,i} (\hat{S}_{tji} - S_{tji})^2 / \sum_{t,j,i} S_{tji}^2
\end{equation}
where $S_{tji}$ is the low-rank signal: $\mathbf{S}_{t}=\mathbf{U} \mathbf{V}^T$.
Table \ref{table: sim_complete} shows the results for all methods. For each covariance structure, the correctly specified model consistently performs the best, as one would reasonably expect. 
When SNR is high, the model can perform well without modeling correlation, even when the true underlying structure is highly correlated. However, the margin of improvement in estimation accuracy is more significant under low SNR settings, demonstrating that the borrowing of information is effective when the data itself has low information. The TEMPTED method does not perform in a comparable range in this experiment as it does not assume a correlated low rank matrix structure. In the misspecified scenario, the \textit{free} covariance model performs consistently the best across SNR ratios and correlation level, since it can be considered as correctly specified but overparametrized, which leads to more robust estimation at the cost of lower efficiency. {\chLB Lastly, we note that in all simulation repetitions, all models selected the true rank.}

\begin{table}[ht]
\centering
\caption{RSE for complete data estimation}
\label{table: sim_complete}
\scriptsize
\begin{tabular}{|l|cc|cc|cc|}
\hline
\multicolumn{7}{|c|}{\textbf{RBF Scenario}} \\
\hline
& \multicolumn{2}{c|}{{\chLB Low SNR}} & \multicolumn{2}{c|}{{\chLB Medium SNR}} & \multicolumn{2}{c|}{{\chLB High SNR}} \\
\hline
Method & Low Corr & High Corr & Low Corr & High Corr & Low Corr & High Corr \\
\hline
EC & 0.239 (0.007) & 0.184 (0.005) & 0.042 (0.001) & 0.043 (0.001) & 0.012 (0.000) & 0.011 (0.000) \\
RBF & \textbf{0.213} (0.006) & \textbf{0.139} (0.005) & \textbf{0.040} (0.001) & \textbf{0.029} (0.001) & \textbf{0.011} (0.000) & \textbf{0.008} (0.000) \\
Free & 0.218 (0.006) & 0.149 (0.005) & 0.040 (0.001) & 0.030 (0.001) & \textbf{0.011} (0.000) & 0.008 (0.000) \\
EVB & 0.251 (0.007) & 0.251 (0.008) & 0.042 (0.001) & 0.047 (0.002) & 0.012 (0.000) & 0.012 (0.000) \\
TEMPTED & 0.509 (0.023) & 0.515 (0.018) & 0.110 (0.002) & 0.080 (0.003) & 0.073 (0.001) & 0.016 (0.001) \\
\hline
\hline
\multicolumn{7}{|c|}{\textbf{EC Scenario}} \\
\hline
EC & \textbf{0.244} (0.007) & \textbf{0.181} (0.003) & \textbf{0.045} (0.001) & \textbf{0.040} (0.001) & \textbf{0.012} (0.000) & \textbf{0.011} (0.000) \\
RBF & 0.248 (0.007) & 0.231 (0.003) & \textbf{0.045} (0.001) & 0.042 (0.001) & \textbf{0.012} (0.000) & \textbf{0.011} (0.000) \\
Free & 0.247 (0.007) & 0.189 (0.003) & \textbf{0.045} (0.001) & \textbf{0.040} (0.001) & \textbf{0.012} (0.000) & \textbf{0.011} (0.000) \\
EVB & 0.254 (0.007) & 0.251 (0.005) & \textbf{0.045} (0.001) & 0.043 (0.001) & \textbf{0.012} (0.000) & 0.012 (0.000) \\
TEMPTED & 0.647 (0.018) & 0.639 (0.017) & 0.247 (0.004) & 0.109 (0.003) & 0.208 (0.003) & 0.062 (0.002) \\
\hline
\hline
\multicolumn{7}{|c|}{\textbf{Free Scenario}} \\
\hline
EC & \multicolumn{2}{c|}{0.243 (0.024)} & \multicolumn{2}{c|}{0.046 (0.005)} & \multicolumn{2}{c|}{0.015 (0.001)} \\
RBF & \multicolumn{2}{c|}{0.242 (0.024)} & \multicolumn{2}{c|}{0.046 (0.005)} & \multicolumn{2}{c|}{0.015 (0.001)} \\
Free & \multicolumn{2}{c|}{\textbf{0.203} (0.022)} & \multicolumn{2}{c|}{\textbf{0.040} (0.005)} & \multicolumn{2}{c|}{\textbf{0.013} (0.001)} \\
EVB & \multicolumn{2}{c|}{0.248 (0.026)} & \multicolumn{2}{c|}{0.047 (0.006)} & \multicolumn{2}{c|}{0.015 (0.001)} \\
TEMPTED & \multicolumn{2}{c|}{0.536 (0.080)} & \multicolumn{2}{c|}{0.089 (0.009)} & \multicolumn{2}{c|}{0.050 (0.006)} \\
\hline
\hline
\multicolumn{7}{|c|}{\textbf{Block Scenario}} \\
\hline
EC & \textbf{0.250} (0.006) & 0.232 (0.004) & \textbf{0.045} (0.001) & 0.044 (0.001) & \textbf{0.012} (0.000) & \textbf{0.011} (0.000) \\
RBF & \textbf{0.250} (0.006) & 0.225 (0.004) & \textbf{0.045} (0.001) & 0.043 (0.001) & \textbf{0.012} (0.000) & \textbf{0.011} (0.000) \\
Free & 0.252 (0.006) & \textbf{0.203} (0.004) & \textbf{0.045} (0.001) & \textbf{0.042} (0.001) & \textbf{0.012} (0.000) & \textbf{0.011} (0.000) \\
EVB & 0.256 (0.006) & 0.245 (0.005) & \textbf{0.045} (0.001) & 0.044 (0.001) & \textbf{0.012} (0.000) & \textbf{0.011} (0.000) \\
TEMPTED & 0.669 (0.018) & 0.551 (0.012) & 0.260 (0.003) & 0.111 (0.002) & 0.227 (0.003) & 0.068 (0.002) \\
\hline
\end{tabular}
\end{table}

\subsection{Simulation study 2: Missing data imputation} \label{sec:sim missing imp}

Here we evaluate the imputation performance of our model. We consider both entrywise and columnwise missingness, where the latter refers to entire observations missing at a timepoint. Data are generated as in Section \ref{sec:complete_sim}, then {\chLB $10\%$} of the data are randomly set to be missing. In addition to the methods in experiment $1$, we compare our model against two imputation algorithms - BAMITA, which is a Bayesian tensor imputation model \citep{BAMITA}, as well as MissForest, a non-parametric, random forest-based imputation method that doesn't assume low rank latent structure \citep{missforest}. We evaluate imputation accuracy by restricting to the $RSE$ for missing entries only.
\begin{equation}
    \text{RSE}_{miss} = \sum_{(t,j,i) \in \mathcal{M}} (\hat{S}_{tji} - S_{tji})^2 / \sum_{(t,j,i) \in \mathcal{M}} S_{tji}^2
\end{equation}
We note that TEMPTED cannot handle entrywise missing data so it is excluded from the setting. The results for entrywise missingness are shown in Table \ref{table: sim_entry_mis}. The patterns closely mirror the complete data estimation results from Table~\ref{table: sim_complete}, where the correctly specified covariance model consistently performs the best, especially in low SNR settings. Alternative imputation methods perform markedly worse across all scenarios, with BAMITA performing substantially worse in high SNR settings, due to the strong signal indicating a non-tensor data structure. MissForest exhibits substantially higher errors potentially due to not assuming low rank structure or an explicit correlation structure across time.

\begin{table}[!htbp]
\centering
\caption{RSE for entrywise missingness imputation}
\label{table: sim_entry_mis}
\scriptsize
\begin{tabular}{|l|cc|cc|cc|}
\hline
\multicolumn{7}{|c|}{\textbf{RBF Scenario}} \\
\hline
& \multicolumn{2}{c|}{{\chLB Low SNR}} & \multicolumn{2}{c|}{{\chLB Medium SNR}} & \multicolumn{2}{c|}{{\chLB High SNR}} \\
\hline
Method & Low Corr & High Corr & Low Corr & High Corr & Low Corr & High Corr \\
\hline
EC & 0.280 (0.010) & 0.212 (0.007) & 0.051 (0.001) & 0.049 (0.002) & \textbf{0.013} (0.000) & 0.013 (0.001) \\
RBF & \textbf{0.249} (0.009) & \textbf{0.161} (0.006) & \textbf{0.048} (0.001) & \textbf{0.033} (0.001) & \textbf{0.013} (0.000) & \textbf{0.009} (0.000) \\
Free & 0.255 (0.009) & 0.187 (0.009) & 0.049 (0.001) & 0.034 (0.001) & \textbf{0.013} (0.000) & \textbf{0.009} (0.000) \\
EVB & 0.295 (0.010) & 0.295 (0.008) & 0.052 (0.001) & 0.054 (0.002) & 0.014 (0.000) & 0.013 (0.001) \\
BAMITA & 0.390 (0.012) & 0.272 (0.009) & 0.115 (0.002) & 0.053 (0.002) & 0.075 (0.001) & 0.013 (0.001) \\
MissForest & 0.802 (0.018) & 0.810 (0.015) & 0.211 (0.004) & 0.223 (0.004) & 0.089 (0.002) & 0.089 (0.003) \\
\hline
\hline
\multicolumn{7}{|c|}{\textbf{EC Scenario}} \\
\hline
& \multicolumn{2}{c|}{{\chLB Low SNR}} & \multicolumn{2}{c|}{{\chLB Medium SNR}} & \multicolumn{2}{c|}{{\chLB High SNR}} \\
\hline
Method & Low Corr & High Corr & Low Corr & High Corr & Low Corr & High Corr \\
\hline
EC & \textbf{0.287} (0.012) & \textbf{0.199} (0.006) & \textbf{0.051} (0.002) & \textbf{0.048} (0.001) & \textbf{0.014} (0.000) & \textbf{0.013} (0.000) \\
RBF & 0.290 (0.013) & 0.246 (0.006) & \textbf{0.051} (0.002) & 0.051 (0.002) & \textbf{0.014} (0.000) & 0.014 (0.000) \\
Free & 0.291 (0.012) & 0.211 (0.007) & \textbf{0.051} (0.002) & \textbf{0.048} (0.002) & \textbf{0.014} (0.000) & \textbf{0.013} (0.000) \\
EVB & 0.296 (0.013) & 0.291 (0.009) & 0.052 (0.002) & 0.053 (0.002) & \textbf{0.014} (0.000) & 0.014 (0.000) \\
BAMITA & 0.528 (0.013) & 0.314 (0.008) & 0.240 (0.004) & 0.103 (0.002) & 0.206 (0.005) & 0.065 (0.002) \\
MissForest & 0.800 (0.019) & 0.804 (0.020) & 0.209 (0.006) & 0.210 (0.006) & 0.088 (0.003) & 0.088 (0.003) \\
\hline
\hline
\multicolumn{7}{|c|}{\textbf{Free Scenario}} \\
\hline
& \multicolumn{2}{c|}{{\chLB Low SNR}} & \multicolumn{2}{c|}{{\chLB Medium SNR}} & \multicolumn{2}{c|}{{\chLB High SNR}} \\
\hline
Method & \multicolumn{6}{c|}{Single Condition} \\
\hline
EC & \multicolumn{2}{c|}{0.332 (0.027)} & \multicolumn{2}{c|}{0.051 (0.004)} & \multicolumn{2}{c|}{0.015 (0.001)} \\
RBF & \multicolumn{2}{c|}{0.331 (0.028)} & \multicolumn{2}{c|}{0.051 (0.004)} & \multicolumn{2}{c|}{0.015 (0.001)} \\
Free & \multicolumn{2}{c|}{\textbf{0.283} (0.040)} & \multicolumn{2}{c|}{\textbf{0.043} (0.004)} & \multicolumn{2}{c|}{\textbf{0.014} (0.001)} \\
EVB & \multicolumn{2}{c|}{0.339 (0.029)} & \multicolumn{2}{c|}{0.051 (0.004)} & \multicolumn{2}{c|}{0.015 (0.001)} \\
BAMITA & \multicolumn{2}{c|}{0.401 (0.040)} & \multicolumn{2}{c|}{0.082 (0.007)} & \multicolumn{2}{c|}{0.051 (0.006)} \\
MissForest & \multicolumn{2}{c|}{0.883 (0.052)} & \multicolumn{2}{c|}{0.215 (0.014)} & \multicolumn{2}{c|}{0.098 (0.005)} \\
\hline
\hline
\multicolumn{7}{|c|}{\textbf{Block Scenario}} \\
\hline
& \multicolumn{2}{c|}{{\chLB Low SNR}} & \multicolumn{2}{c|}{{\chLB Medium SNR}} & \multicolumn{2}{c|}{{\chLB High SNR}} \\
\hline
Method & Low Corr & High Corr & Low Corr & High Corr & Low Corr & High Corr \\
\hline
EC & \textbf{0.286} (0.005) & 0.277 (0.007) & \textbf{0.051} (0.001) & 0.054 (0.001) & \textbf{0.013} (0.000) & 0.014 (0.000) \\
RBF & 0.287 (0.005) & 0.266 (0.006) & \textbf{0.051} (0.001) & 0.053 (0.001) & \textbf{0.013} (0.000) & 0.014 (0.000) \\
Free & 0.288 (0.005) & \textbf{0.245} (0.006) & \textbf{0.051} (0.002) & \textbf{0.050} (0.001) & \textbf{0.013} (0.000) & \textbf{0.013} (0.000) \\
EVB & 0.291 (0.005) & 0.295 (0.008) & \textbf{0.051} (0.001) & 0.054 (0.001) & \textbf{0.013} (0.000) & 0.014 (0.000) \\
BAMITA & 0.517 (0.007) & 0.363 (0.009) & 0.255 (0.005) & 0.111 (0.002) & 0.212 (0.004) & 0.066 (0.001) \\
MissForest & 0.795 (0.010) & 0.800 (0.014) & 0.209 (0.005) & 0.222 (0.004) & 0.086 (0.003) & 0.090 (0.002) \\
\hline
\end{tabular}
\end{table}

Table \ref{table: sim_block_mis} summarizes the results for columnwise missingness imputation. The task becomes much more difficult, as imputation completely relies on shared information between timepoints. This is shown by the results of low correlation scenarios under both the $EC$ and $RBF$ correlation structure, where RSE vary little over SNR levels. Even for the correct model, RSE is significantly worse than in estimation or entrywise imputation. All alternative methods other than BAMITA have RSE$>1$, which is the RSE of null imputing with $0$. It is also worth noting that misspecification of the correlation structure is more costly, as shown by the relatively worse performance of our model in misspecified scenarios compared to estimation and entrywise imputation. Only the most flexible free covariance could still achieve an RSE well below $1$. On the other hand, it is remarkable that the proposed method performs relatively well even under the lowest SNR under high correlation, recovering about 70\% of true structure in both the EC and RBF scenarios. {\chLB In all repetitions, all models selected the true rank.}

\begin{table}[!htbp]
\centering
\caption{RSE of blockwise imputation for all methods under varying simulated conditions, with the best performing method highlighted in bold font.}
\label{table: sim_block_mis}
\scriptsize
\begin{tabular}{|l|cc|cc|cc|}
\hline
\multicolumn{7}{|c|}{\textbf{RBF Scenario}} \\
\hline
& \multicolumn{2}{c|}{{\chLB Low SNR}} & \multicolumn{2}{c|}{{\chLB Medium SNR}} & \multicolumn{2}{c|}{{\chLB High SNR}} \\
\hline
Method & Low Corr & High Corr & Low Corr & High Corr & Low Corr & High Corr \\
\hline
EC & 0.880 (0.071) & 0.352 (0.037) & 0.843 (0.057) & 0.291 (0.052) & 0.847 (0.022) & 0.282 (0.038) \\
RBF & \textbf{0.627} (0.061) & \textbf{0.234} (0.033) & \textbf{0.462} (0.050) & \textbf{0.083} (0.014) & \textbf{0.449} (0.063) & \textbf{0.042} (0.009) \\
Free & 0.670 (0.066) & 0.264 (0.042) & 0.476 (0.050) & 0.089 (0.015) & 0.462 (0.068) & 0.044 (0.008) \\
EVB & 1.000 (0.000) & 1.000 (0.000) & 1.000 (0.000) & 1.000 (0.000) & 1.000 (0.000) & 1.000 (0.000) \\
BAMITA & 0.874 (0.082) & 0.350 (0.061) & 0.652 (0.105) & 0.124 (0.021) & 0.900 (0.233) & 0.059 (0.013) \\
MissForest & 11.205 (1.150) & 10.368 (1.577) & 3.430 (0.432) & 3.392 (0.346) & 2.443 (0.319) & 2.279 (0.219) \\
TEMPTED & 2.005 (0.401) & 1.978 (0.132) & 1.643 (0.204) & 1.776 (0.197) & 1.508 (0.139) & 1.908 (0.300) \\
\hline
\hline
\multicolumn{7}{|c|}{\textbf{EC Scenario}} \\
\hline
EC & \textbf{0.928} (0.030) & \textbf{0.335} (0.027) & \textbf{0.931} (0.036) & \textbf{0.278} (0.033) & \textbf{0.916} (0.018) & \textbf{0.254} (0.027) \\
RBF & 0.976 (0.018) & 0.380 (0.048) & 0.973 (0.028) & 0.557 (0.064) & 0.970 (0.020) & 0.506 (0.050) \\
Free & 0.964 (0.028) & 0.373 (0.028) & 0.975 (0.064) & 0.294 (0.039) & 0.932 (0.025) & 0.263 (0.032) \\
EVB & 1.000 (0.000) & 1.000 (0.000) & 1.000 (0.000) & 1.000 (0.000) & 1.000 (0.000) & 1.000 (0.000) \\
BAMITA & 1.325 (0.163) & 0.451 (0.033) & 2.723 (0.255) & 0.516 (0.069) & 5.835 (1.340) & 0.812 (0.220) \\
MissForest & 9.877 (1.417) & 10.730 (1.043) & 3.556 (0.396) & 3.282 (0.224) & 2.459 (0.164) & 2.249 (0.213) \\
TEMPTED & 1.723 (0.216) & 2.172 (0.317) & 1.519 (0.140) & 1.811 (0.188) & 1.500 (0.162) & 1.856 (0.273) \\
\hline
\hline
\multicolumn{7}{|c|}{\textbf{Free Scenario}} \\
\hline
EC & \multicolumn{2}{c|}{0.999 (0.014)} & \multicolumn{2}{c|}{0.970 (0.042)} & \multicolumn{2}{c|}{0.986 (0.032)} \\
RBF & \multicolumn{2}{c|}{0.981 (0.037)} & \multicolumn{2}{c|}{0.971 (0.035)} & \multicolumn{2}{c|}{0.976 (0.040)} \\
Free & \multicolumn{2}{c|}{\textbf{0.662} (0.196)} & \multicolumn{2}{c|}{\textbf{0.364} (0.126)} & \multicolumn{2}{c|}{\textbf{0.318} (0.133)} \\
EVB & \multicolumn{2}{c|}{1.000 (0.000)} & \multicolumn{2}{c|}{1.000 (0.000)} & \multicolumn{2}{c|}{1.000 (0.000)} \\
BAMITA & \multicolumn{2}{c|}{0.807 (0.222)} & \multicolumn{2}{c|}{0.477 (0.147)} & \multicolumn{2}{c|}{0.673 (0.415)} \\
MissForest & \multicolumn{2}{c|}{11.321 (2.702)} & \multicolumn{2}{c|}{3.653 (0.504)} & \multicolumn{2}{c|}{2.353 (0.165)} \\
TEMPTED & \multicolumn{2}{c|}{2.041 (0.358)} & \multicolumn{2}{c|}{1.793 (0.183)} & \multicolumn{2}{c|}{1.784 (0.234)} \\
\hline
\hline
\multicolumn{7}{|c|}{\textbf{Block Scenario}} \\
\hline
EC & \textbf{0.979} (0.029) & 0.866 (0.071) & 0.981 (0.018) & 0.843 (0.054) & 0.991 (0.020) & 0.868 (0.028) \\
RBF & 0.983 (0.020) & 0.697 (0.063) & \textbf{0.972} (0.023) & 0.651 (0.058) & 0.988 (0.015) & 0.690 (0.044) \\
Free & 0.999 (0.045) & \textbf{0.549} (0.041) & 0.990 (0.047) & \textbf{0.405} (0.040) & \textbf{0.969} (0.030) & \textbf{0.418} (0.056) \\
EVB & 1.000 (0.000) & 1.000 (0.000) & 1.000 (0.000) & 1.000 (0.000) & 1.000 (0.000) & 1.000 (0.000) \\
BAMITA & 1.416 (0.123) & 0.700 (0.053) & 3.010 (0.381) & 0.633 (0.071) & 5.545 (1.837) & 0.985 (0.125) \\
MissForest & 10.182 (0.997) & 11.127 (1.703) & 3.275 (0.332) & 3.389 (0.325) & 2.384 (0.167) & 2.209 (0.134) \\
TEMPTED & 1.801 (0.163) & 2.093 (0.332) & 1.616 (0.155) & 1.686 (0.241) & 1.564 (0.112) & 26.49 (77.14) \\

\hline
\end{tabular}
\end{table}

\section{Gene Time Course Data Application}
\label{Sec: Real Data}

\begin{figure}[!ht]
 \centering \includegraphics[width=1\linewidth]{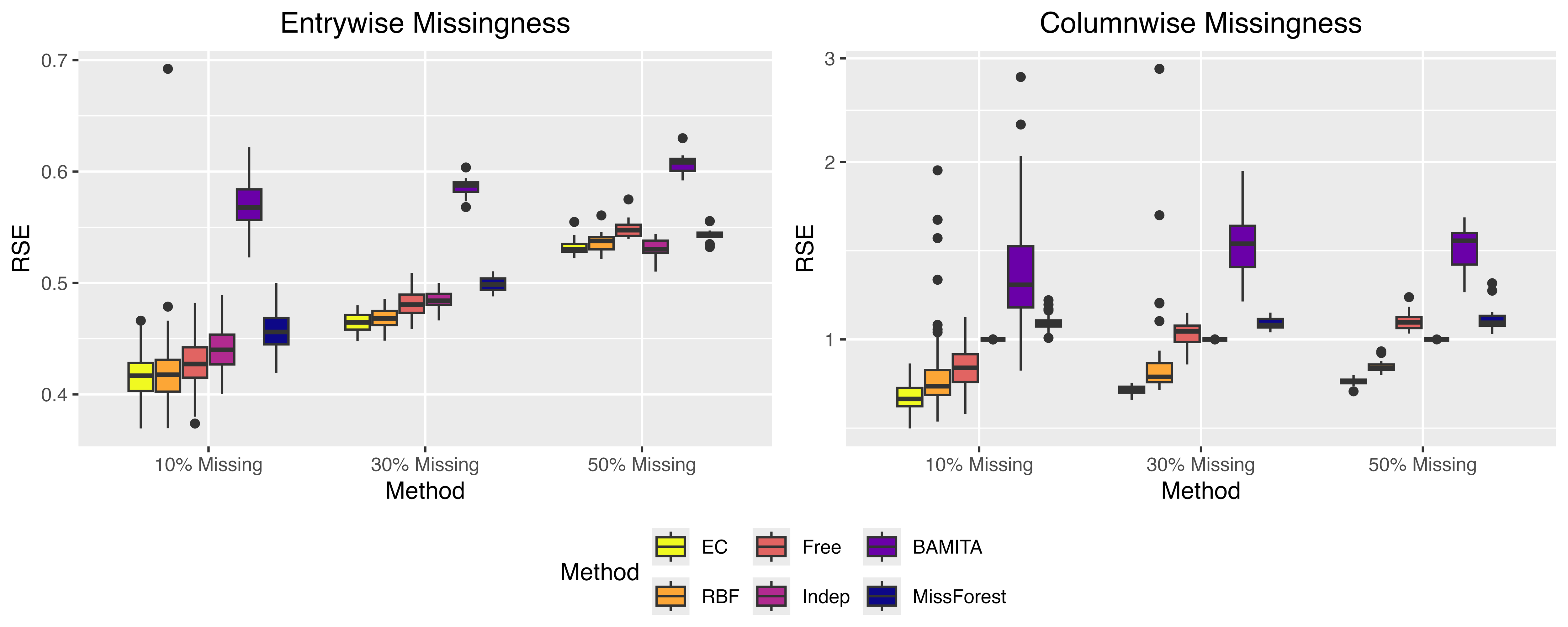}
  \caption{Analysis results for the IFN$\beta$ data application: Imputation accuracy for all methods from cross validation.}
\label{fig:ifnb_miss_sub}
\end{figure}

We apply our method to a set of gene time course data described in \cite{IFNB}. This study explored methods to classify genetic response of multiple sclerosis patients to recombinant human interferon beta(rIFN$\beta$) treatment. rIFN$\beta$ was given to $53$ patients and expression levels of $76$ genes were measured at $7$ timepoints (baseline and follow-ups at $3,6,9,12,18,24$ months after treatment). Most patients had some percentage of missingness, ranging from $1-6$ entire timepoints missing and additional partial missingness at observed timepoints. Overall, $10.5\%$ of all data were missing. We first use this dataset to evaluate the empirical imputation performance of our model. Similar to the experiments in section \ref{sec:sim missing imp}, we apply our model with the three non-trival covariance functions, EVB under independence, BAMITA, and MissForest to the data. TEMPTED is omitted as it cannot handle entrywise missingness. We take the complete indices and perform $10,3,2$-fold cross validation based on the missing proportion, repeated $10$ times over different random seeds. Figure \ref{fig:ifnb_miss_sub} shows the results. 

\begin{figure}[!ht]
    \centering
    \subfloat[Variance explained and {\chLB estimated correlation ($\hat{\rho}$)} for the first 7 components. \label{fig:ifnb_var_exp_sub}]{%
        \includegraphics[width=0.60\linewidth]{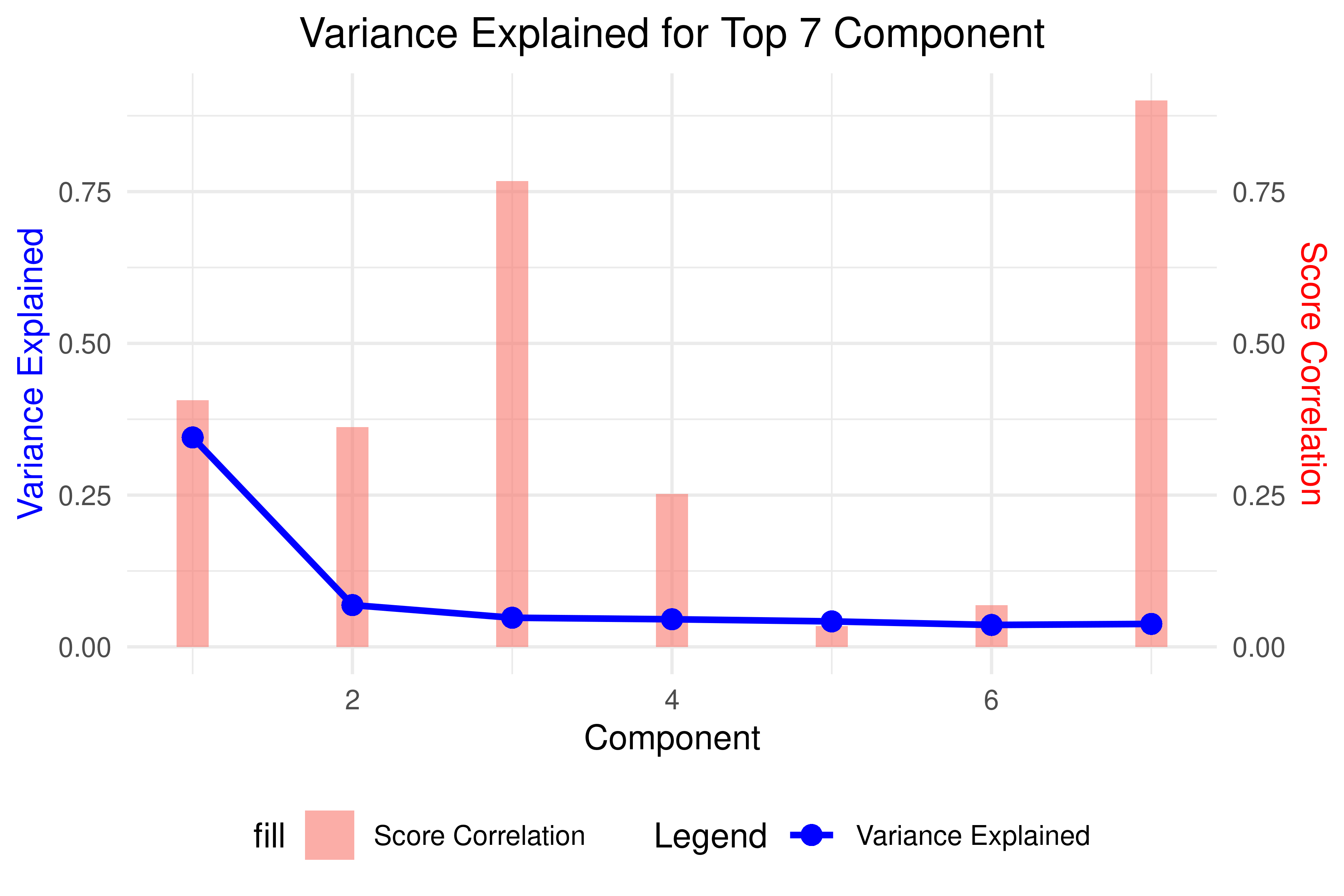}
    }
    \vspace{0.3cm}
    \subfloat[Individual trajectories for the first and seventh component scores. \label{fig:ifnb_score_plot_sub}]{%
        \includegraphics[width=0.85\linewidth]{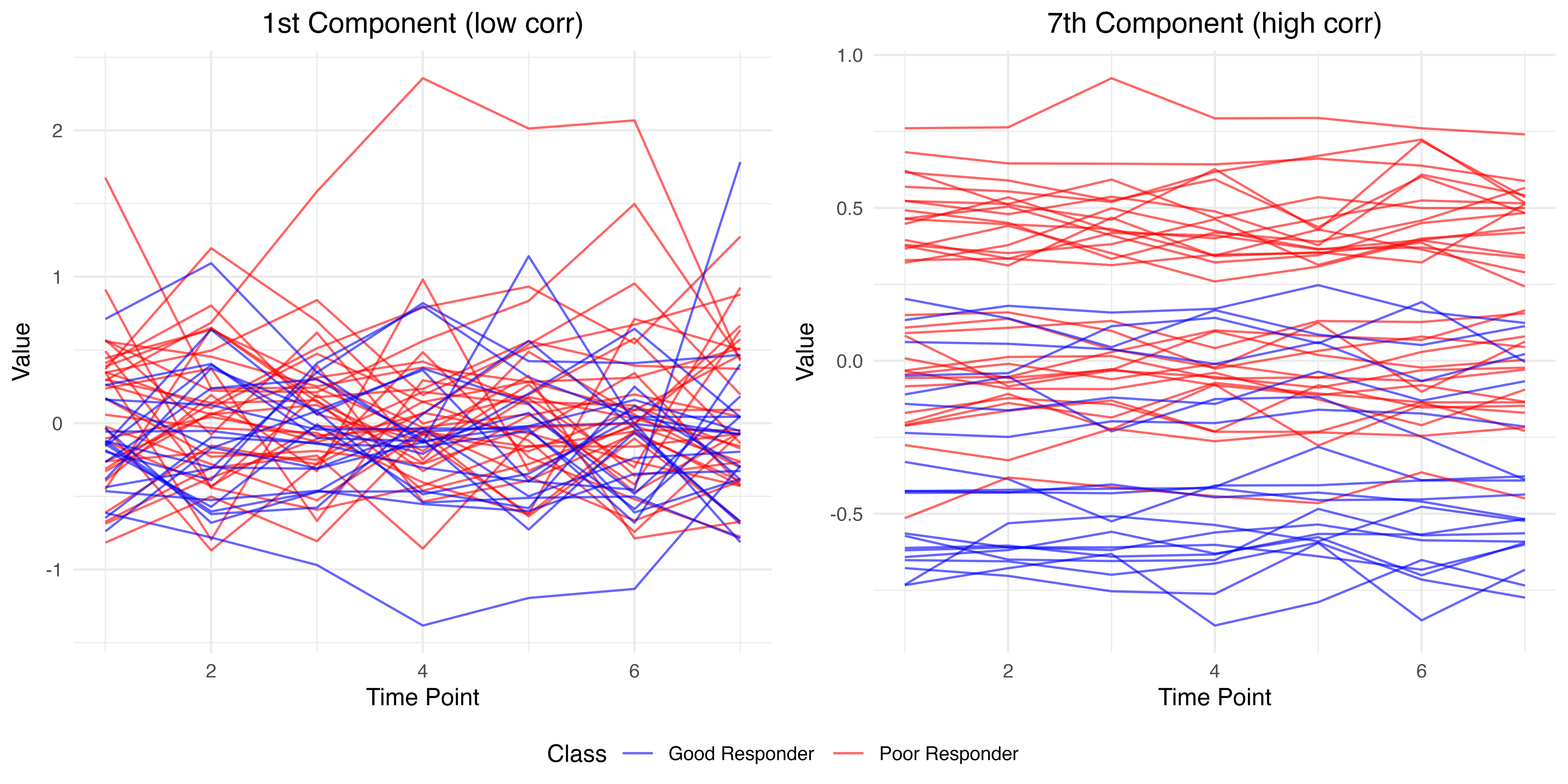}
    }
    \caption{Analysis results for the IFN$\beta$ data application.}
    \label{fig:combined_ifnb}
\end{figure}

In the comparisons, the \textit{EC} covariance model performed the best across both entrywise and columnwise missingness. The independent model performed comparably in the entrywise missing scenarios, but had $RSE=1$ in columnwise missingness, consistent with simulation results. All alternative methods performed worse than the independent model, highlighting the advantage of our proposed model in columnwise imputation.

As the \textit{EC} covariance model performed the best, we refer to the results produced by this model for interpretation. %
We estimated a total of $14$ components; the top $7$ ordered by variance explained are shown in Figure \ref{fig:ifnb_var_exp_sub}, along with the estimated correlation in each component's scores. The first component explains about $35\%$ variance, while the next $6$ each explains less than $10\%$ variance each. Both the third and seventh component exhibits high correlation between sample scores at different timepoints. We plot individual score trajectories in Figure \ref{fig:ifnb_score_plot_sub}, colored by manually labeled good/poor responders from the original study. The first component, which is the major contributor, exhibits similar patterns for most individuals, indicating a mean effect of subjects. The seventh component, on the other hand, shows a clear clustering effect in good and poor responders, with good responders at the bottom and poor responders at the top. The individual scores have little variation between time points, which indicates that the two types of responders can be distinguished without inferring from temporal patterns. This is consistent with the results in \cite{IFNB}, where no group*time interaction effects were discovered; as well as \cite{msDWD}, where distinction between the two groups were not driven by factors that vary over time.
More generally, the EC covariance and accompanying score trajectories suggest some clear subject effects but no other temporal trends. 

\begin{figure}[!ht]
 \centering \includegraphics[width=1\linewidth]{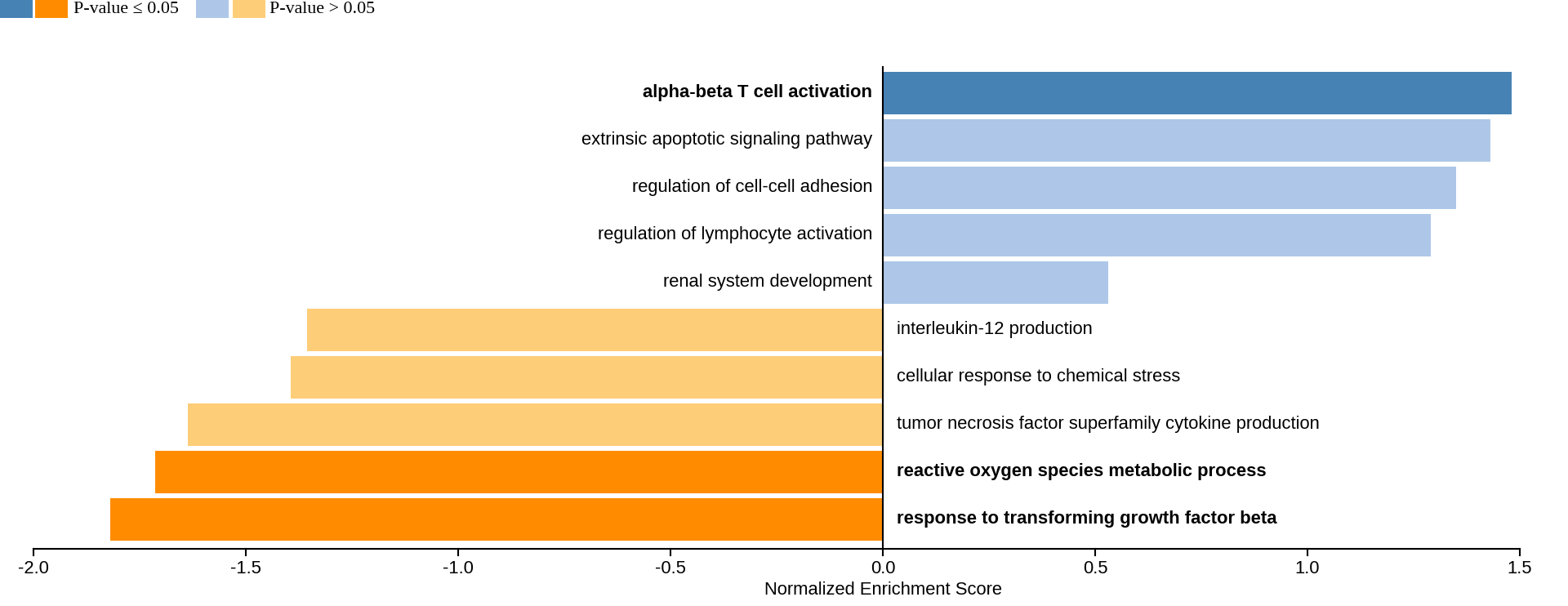}
  \caption{{\chLB Gene set enrichment analysis for loadings of Component 7 in the IFN$\beta$ factorization}}
\label{fig:ifnb_comp7_enrichment}
\end{figure}

{\chLB To further investigate the biological relevance of Component 7, we performed a gene-set enrichment analysis of its loadings using GO Biological Processes via WebGestalt \cite{elizarraras2024webgestalt}.  While no processes were significant at the FDR $<0.05$ level, the top enriched processes and their nominal significance is shown in Figure~\ref{fig:ifnb_comp7_enrichment}.  The most significant and top downregulated process, implying higher levels for good responders, was response to transforming growth factor beta; this is consistent with prior literature associating higher TGF-$\beta$ levels with better treatment response \cite{cucci2010pro}.}

\section{Discussion}
\label{Sec: Discussion}

In this paper, we introduce a latent matrix factorization method for high dimensional longitudinal data that captures temporal correlation by flexibly modeling the covariance structure in latent factor scores. The additive nature of our model enables decomposition of data into static and temporal factors, enhancing interpretability. The model employs a computationally efficient empirical variational Bayes estimation algorithm that automatically estimates hyperparameters and selects model dimension by optimizing an objective function balancing model fit and complexity. Through extensive simulation studies and a real world data application, the proposed method demonstrates superior estimation and imputation accuracy compared to existing approaches across diverse scenarios.

While we focus on single-source data, multimodal datasets combining data from multiple sites are increasingly common in biomedical studies, such as collaborative neurodegenerative disease research that collects imaging, clinical, and genetic modalities \cite{TRACKFA_cross_sec,Oh_etal_AD,TRACKHD}. Extensions of our method to multimodal matrix factorization—where some components are restricted to specific cohorts or data modalities—is an interesting direction for future research, as the full rank $R$ and greedy estimation algorithms do not straightforwardly extend to such settings.

A known limitation of variational Bayes methods is uncertainty quantification. While point estimates generally perform well and have good properties, variational Bayes methods typically underestimate marginal posterior variances \cite{Wang2004_VB,Turner2011_VB_Timeseries}. To address this limitation, our method can serve as initialization for the associated fully Bayesian model, offering several advantages including initialization near the posterior mode and automatic rank selection, which are challenging for the fully Bayesian approach to achieve alone. A related Bayesian model is described in \cite{Samorodnitsky2024}, which illustrates the benefits of accounting for uncertainty in imputation and prediction. However, standard MCMC strategies such as Gibbs sampling would likely face substantial computational challenges, necessitating the development of more efficient sampling algorithms. {\chLB As our model imputes values given the fit to observed data, it implicitly assumes data are missing at random; extensions to  scenarios where the missing mechanism is not random but known (i.e., a limit of detection) is another direction of future work.}

\section*{Code \& Data Availability}

Documented R code to perform the proposed method, {\chLB including an R package}, and the IFN$\beta$ dataset analyzed in Section \ref{Sec: Real Data} is available at \\
\noindent \url{https://github.com/andyljk/Longitudinal_Factorization}. \\
{\chLB Additionally, a supplementary zipped folder includes code to perform all the simulations and data analysis in the manuscript.}

\section*{Supplementary Materials}

{\chLB A supplementary document available online includes proofs, details for the method with irregular timepoints, several additional simulation studies, additional computational details, and further details for the real data application.}

\bibliographystyle{unsrtnat}
\bibliography{references}

@article{EBMF,
    author = {Wang, W. and Stephens, M.},
    title = {Empirical {B}ayes Matrix Factorization},
    journal = {Journal of Machine Learning Research},
    volume = {22(120)},
    number = {},
    pages = {1--40},
    year = {2021}
}

@article{EBMNM,
    author = "Yang, Y. and Carbonetto, P. and Gerard, D. and Stephens, M.",
    title = "{Improved Methods for Empirical {B}ayes Multivariate Multiple Testing and Effect Size Estimation}.",
    journal = "arXiv preprint arXiv: 2406.08784",
    year = "2024"
}

@article{VI_Review,
    author = {Blei, D. M. and Kucukelbir, A. and McAuliffe, J. D.},
    title = {Variational Inference: a review for statisticians},
    journal = {Journal of the American Statistical Association},
    volume = {112},
    number = {518},
    pages = {859--877},
    year = {2017},
}

@article{MCMC,
    author = {Neal, R. M.},
    title = {Probabilistic Inference Using {M}arkov Chain {M}onte {C}arlo Methods},
    journal = {Technical Reports},
    volume = {Department of Computer Science},
    number = {},
    pages = {University of Toronto},
    year = {1993},
}

@article{EV_BIDI,
    author = {Lock, E. F.},
    title = {Empirical {B}ayes linked matrix decomposition},
    journal = {Machine Learning},
    volume = {113},
    number = {},
    pages = {7451--7477},
    year = {2024},
}

@article{BAMITA,
    author = {Jiang, Z. and Li, G. and Lock, E. F.},
    title = {{BAMITA}: {B}ayesian multiple imputation for tensor arrays},
    journal = {Biostatistics},
    volume = {26},
    number = {1},
    pages = {},
    year = {2025}
}

@article{nakajima13a,
  author  = {Nakajima, S. and Sugiyama, M. and Babacan, S. D. and Tomioka, R.},
  title   = {Global Analytic Solution of Fully-observed Variational {B}ayesian Matrix Factorization},
  journal = {Journal of Machine Learning Research},
  year    = {2013},
  volume  = {14},
  number  = {1},
  pages   = {1--37}
}

@article{TEMPTED,
    author = {Shi, P. and Martino, C. and Han, R. and et al.},
    title = {{TEMPTED}: time-informed dimensionality reduction for longitudinal microbiome studies.},
    journal = {Genome Biology},
    volume = {25},
    number = {317},
    pages = {891--921},
    year = {2024},
    DOI = {https://doi.org/10.1186/s13059-024-03453-x}
}

@article{missforest,
    author = {Stekhoven, D. J. and Bühlmann, P},
    title = {Missforest: non-parametric missing value
imputation for mixed-type data},
    journal = {Bioinformatics},
    volume = {28},
    number = {1},
    pages = {112--128},
    year = {2012},
    DOI = {}
}

@article{IFNB,
    author = {Baranzini, S. E. and Mousavi, P. and Rio, J. and Caillier, S. J. and Stillman, A. and Villoslada, P. and Wyatt, M. M. and Comabella, M. and Greller, L. D. and Somogyi, R. and Montalban,X. and Oksenberg, J. O.},
    title = {Transcription-based Prediction of Response to {IFN}$\beta$ using Supervised Computational Methods},
    journal = {PloS Biology},
    volume = {3},
    number = {1},
    pages = {e2},
    year = {2004},
    DOI = {}
}

@article{msDWD,
    author = {Guo, B. and Eberly, L. E. and Henry, P. G. and Lenglet, C. and Lock, E. F.},
    title = {Multiway Sparse Distance Weighted Discrimination},
    journal = {Journal of Computational and Graphical Statistics},
    volume = {32},
    number = {2},
    pages = {730--743},
    year = {2022},
    DOI = {https://doi.org/10.1080/10618600.2022.2099404}
}

@article{softimpute,
    author = {Mazumder, R. and Hastie, T. and Tibshirani, R.},
    title = {Spectral Regularization Algorithms for Learning Large Incomplete Matrices},
    journal = {Journal of Machine Learning Research},
    volume = {11},
    number = {80},
    pages = {2287--2322},
    year = {2010}
}

@article{MFAI,
    author = {Wang, Z. and Zhang, F. and Zheng, C. and Hu, X. and Cai, M. and Yang, C.},
    title = { {MFAI} : A Scalable {B}ayesian Matrix Factorization Approach to Leveraging Auxiliary Information},
    journal = {Journal of Computational and Graphical Statistics},
    volume = {33},
    number = {4},
    pages = {1339--1349},
    year = {2024},
    DOI = {https://doi.org/10.1080/10618600.2024.2319160}
}

@article{Variational_PC,
    author = {Bishop, C. M.},
    title = {Variational principal components},
    journal = {In Ninth International Conference on Artificial Neural Networks},
    volume = {1},
    number = {470},
    pages = {509--514},
    year = {1999}
}

@article{latent_factor_lmm,
    author = {An, X. and Yang, Q. and Bentler, P.M. },
    title = {A latent factor linear mixed model for high-dimensional longitudinal data analysis},
    journal = {Statistics in Medicine},
    volume = {32},
    number = {24},
    pages = {4229--4239},
    year = {2013}
}

@article{bayes_factor_lu_etal,
    author = {Lu, Z. and Chandra, N.K.},
    title = {A sparse factor model for clustering high-dimensional longitudinal data},
    journal = {Statistics in Medicine},
    volume = {43},
    number = {19},
    pages = {3633--3648},
    year = {2024}
}

@article{kolda2009tensor,
  title={Tensor decompositions and applications},
  author={Kolda, Tamara G and Bader, Brett W},
  journal={SIAM review},
  volume={51},
  number={3},
  pages={455--500},
  year={2009},
  publisher={SIAM}
}

@article{guan2024smooth,
  title={Smooth and probabilistic {PARAFAC} model with auxiliary covariates},
  author={Guan, Leying},
  journal={Journal of Computational and Graphical Statistics},
  volume={33},
  number={2},
  pages={538--550},
  year={2024},
  publisher={Taylor \& Francis}
}

@article{fox2012tutorial,
  title={A tutorial on variational {B}ayesian inference},
  author={Fox, Charles W and Roberts, Stephen J},
  journal={Artificial intelligence review},
  volume={38},
  number={2},
  pages={85--95},
  year={2012},
  publisher={Springer}
}

@article{TRACKFA_cross_sec,
  title={Neuroimaging Biomarkers for Friedreich Ataxia: A Cross-Sectional Analysis of the TRACK-FA Study},
  author={Georgiou-Karistianis, Nellie and Corben, Louise A. and Lock, Eric F. and Bujalka, Helena and Adanyeguh, Isaac and Corti, Manuela and Deelchand, Dinesh K. and Delatycki, Martin B. and Dogan, Imis and Farmer, Jennifer and França Jr., Marcondes C. and Gabay, Anthony S. and Gaetz, William and Harding, Ian H. and Joers, James and Lax, Michelle A. and Li, Jiakun and Lynch, David R. and Mareci, Thomas H. and Martinez, Alberto R. M. and Pandolfo, Massimo and Papoutsi, Marina and Parker, Richard G. and Reetz, Kathrin and Rezende, Thiago J.R. and Roberts, Timothy P. and Romanzetti, Sandro and Rudko, David A. and Saha, Susmita and Schulz, Jörg B. and Subramony, Sub H. and Supramaniam, Veena G. and Lenglet, Christophe and Henry, Pierre-Gilles},
  journal={Annals of Neurology},
  volume={98},
  number={2},
  pages={386--397},
  year={2025},
  publisher={Wiley}
}

@article{TRACKHD,
  title={Biological and clinical changes in premanifest and early stage Huntington's disease in the TRACK-HD study: the 12-month longitudinal analysis},
  author={Sarah J Tabrizi and Rachael I Scahill and Alexandra Durr and Raymund AC Roos and Blair R Leavitt and Rebecca Jones and G Bernhard Landwehrmeyer and Nick C Fox and Hans Johnson and Stephen L Hicks and Christopher Kennard and David Craufurd and Chris Frost and Douglas R Langbehn and Ralf Reilmann and Julie C Stout},
  journal={The Lancet Neurology},
  volume={10},
  number={1},
  pages={31--42},
  year={2011}
}

@article{Oh_etal_AD,
  title={A cerebrospinal fluid synaptic protein biomarker for prediction of cognitive resilience versus decline in Alzheimer’s disease.},
  author={Oh, H.SH. and Urey, D.Y. and Karlsson, L. et al.},
  journal={Nature Medicine},
  volume={31},
  number={},
  pages={1592--1603},
  year={2025}
}

@article{Wang2004_VB,
  title={Inadequacy of interval estimates corresponding to variational Bayesian approximations.},
  author={Wang, B and Titterington, M},
  journal={Workshop on Artificial Intelligence and Statistics},
  volume={},
  number={},
  pages={373--380},
  year={2005}
}

@article{Turner2011_VB_Timeseries,
author = {Turner, Richard and Sahani, M. and Barber, D. and Cemgil, Ali and Chiappa, S.},
year = {2011},
pages = {109-130},
title = {Two problems with variational expectation maximisation for time-series models},
journal = {Bayesian Time Series Models}
}

@article{Samorodnitsky2024,
author = {Sarah Samorodnitsky and Chris H. Wendt and Eric F. Lock},
year = {2024},
volume = {197},
title = {Bayesian simultaneous factorization and prediction using multi-omic data},
journal = {Computational Statistics \& Data Analysis}
}

@article{wang2024multilayer,
  title={Multilayer Exponential Family Factor models for integrative analysis and learning disease progression},
  author={Wang, Qinxia and Wang, Yuanjia},
  journal={Biostatistics},
  volume={25},
  number={1},
  pages={203--219},
  year={2024},
  publisher={Oxford University Press}
}

@article{elizarraras2024webgestalt,
  title={WebGestalt 2024: faster gene set analysis and new support for metabolomics and multi-omics},
  author={Elizarraras, John M and Liao, Yuxing and Shi, Zhiao and Zhu, Qian and Pico, Alexander R and Zhang, Bing},
  journal={Nucleic acids research},
  volume={52},
  number={W1},
  pages={W415--W421},
  year={2024},
  publisher={Oxford University Press}
}

@article{cucci2010pro,
  title={Pro-inflammatory cytokine and chemokine mRNA blood level in multiple sclerosis is related to treatment response and interferon-beta dose},
  author={Cucci, Angele and Barbero, Pierangelo and Clerico, Marinella and Ferrero, Bruno and Versino, Elisabetta and Contessa, Giulia and Demercanti, Stefania and Viglietta, Emanuela and Di Liberto, Alessandra and Vai, Alessandra Giai and others},
  journal={Journal of Neuroimmunology},
  volume={226},
  number={1-2},
  pages={150--157},
  year={2010},
  publisher={Elsevier}
}

@article{Hoyer2004,
    author = {Hoyer, Patrik O.},
    title = {Non-negative Matrix Factorization with Sparseness Constraints},
    journal = {Journal of Machine Learning Research},
    volume = {5},
    pages = {1457--1469},
    year = {2004},
    url = {https://www.jmlr.org/papers/v5/hoyer04a.html}
}

@article{gonen2014kernelized,
  author={Gonen, Mehmet and Kaski, Samuel},
  journal={IEEE Transactions on Pattern Analysis and Machine Intelligence}, 
  title={Kernelized {B}ayesian Matrix Factorization}, 
  year={2014},
  volume={36},
  number={10},
  pages={2047-2060}
 }

@inproceedings{singh2008relational,
  title={Relational learning via collective matrix factorization},
  author={Singh, Ajit P and Gordon, Geoffrey J},
  booktitle={Proceedings of the 14th ACM SIGKDD international conference on Knowledge discovery and data mining},
  pages={650--658},
  year={2008}
}

@inproceedings{Bishop_BPCA,
    author = {Bishop, C. M.},
    title = {Bayesian {PCA}},
    booktitle = {Advances in Neural Information Processing Systems 11},
    pages = {382--388},
    publisher = {MIT Press},
    year = {1999}
}

@article{zhao2015bayesian,
    author = {Zhao, Q. and Zhang, L. and Cichocki, A.},
    title = {Bayesian {CP} Factorization of Incomplete Tensors with Automatic Rank Determination},
    journal = {IEEE Transactions on Pattern Analysis and Machine Intelligence},
    volume = {37},
    number = {9},
    pages = {1751--1763},
    year = {2015},
    doi = {10.1109/TPAMI.2015.2392756}
}

@article{nakajima2015condition,
    author = {Nakajima, S. and Tomioka, R. and Sugiyama, M. and Babacan, S. D.},
    title = {Condition for Perfect Dimensionality Recovery by Variational {B}ayesian {PCA}},
    journal = {Journal of Machine Learning Research},
    volume = {16},
    number = {114},
    pages = {3757--3811},
    year = {2015}
}

\clearpage
\setcounter{section}{0}
\renewcommand{\thesection}{S\arabic{section}}
\renewcommand{\theHsection}{supp.\arabic{section}}
\setcounter{equation}{0}
\renewcommand{\theequation}{S\arabic{equation}}
\renewcommand{\theHequation}{supp.\arabic{equation}}
\setcounter{figure}{0}
\renewcommand{\thefigure}{S\arabic{figure}}
\renewcommand{\theHfigure}{supp.\arabic{figure}}
\setcounter{table}{0}
\renewcommand{\thetable}{S\arabic{table}}
\renewcommand{\theHtable}{supp.\arabic{table}}
\phantomsection
\addcontentsline{toc}{part}{Supplementary Materials}
\begin{center}
{\Large Supplementary Materials for ``Empirical Variational Bayes Factorization for Longitudinal Matrix Data''\par}
\medskip
Jiakun Li and Eric F. Lock
\end{center}

\section{Irregular Timepoint Model} \label{Sec: Irreg Model}

In this section, we provide a description of the key differences of the extended model that handles inconsistency in measured timepoints across individuals. For the $i = 1, \dots, n$ individual, assume each with $p$ variables measured at time $t_{1}^{(i)}, \dots, t_{d_i}^{(i)}$. The outcome for individual $i$, variable $j$ at time $t$ is modeled as

\begin{equation}
    \begin{aligned}
        \bm Y_{tji} = \sum_{r=1}^R \bm u_{rj} \bm v_{tri} + \bm E_{tji}
    \end{aligned}
\end{equation}
Note that the matrix product representation in \eqref{equn: model def} is not available in this setting, since $\bm v$ does not have consistent measures at each timepoint. Priors are prescribed to each of $\bm u, \bm v$ similar to \eqref{equn: prior}, where

\begin{equation}
    \begin{aligned}
        \bm u_{rj} \overset{iid}{\sim} N(0,\eta_r^2), \;
        \tilde{\bm{v}}_{ri} \overset{iid}{\sim} mvN(0,\bm K_r^{(i)}), \;
        e_{tji} \overset{iid}{\sim} N(0,\sigma_{tji}^2)
    \end{aligned}
\end{equation}
and $\tilde{\bm v}_{ri} = [\bm v_{t_1^{(i)}ri} \cdots \bm v_{t_{d_i}^{(i)}ri}] \in \mathbb{R}^{d_i}$. The prior covariance matrix $\bm K_r^{(i)} \in \mathbb{R}^{d_i \times d_i}$ has dimension matching the number of measurements from individual $i$. We assume all prior covariance are parametrized by a common set of parameters as described in Section \ref{sec: covariance}, and denote estimates of $\bm K$ as $\hat{\bm K}$ for convenience. When the measurements are highly irregular, modeling the noise variance separately at each timepoints becomes difficult. In this situation, a reasonable modification is to set $\sigma_t^2 = \sigma^2$ for all $t$.

Denoting $\bm U = \{\bm u_{jr}: j=1,\dots, p; r = 1,\dots, R \}$ and $\bm V = \{\bm v_{tir}: t \in T; i=1,\dots, n; r = 1,\dots, R \}$, we maximize the same objective function $F_{ELBO}(q,\bm \eta, \bm K, \sigma)$ using algorithm \ref{alg:rank_R}. For the rank 1 model, it can be shown following the proof of Theorem \ref{thrm EB VB equiv}, step $4,5$ of algorithm \ref{alg:rank_1} are equivalent to solving the following

\begin{equation}
    \begin{aligned}
        \underset{q_{\bm u}, \eta}{\arg \max} \: F_{ELBO}(q_{\bm{u}}, q_{\bm{v}}, \eta, \bm K, \bm{\sigma}) = \text{EBRPM}(\bm{x}_j &= \frac{\sum_{i=1}^n \sum_{t = t_1^{(i)}}^{t_{d_i}^{(i)}} Y_{tji} \overline{v}_{ti}/\sigma^2_t}{\sum_{i=1}^n \sum_{t = t_1^{(i)}}^{t_{d_i}^{(i)}} \overline{v_{ti}^2}/\sigma^2_t}, \\ 
        \bm S_j &= \frac{1}{\sum_{i=1}^n \sum_{t = t_1^{(i)}}^{t_{d_i}^{(i)}} \overline{v_{ti}^2}/\sigma^2_t})
    \end{aligned}
\end{equation}
\textit{and}
\begin{equation}
    \underset{q_{\bm v}, \bm K}{\arg \max} \: F_{ELBO}(q_{\bm{u}}, q_{\bm{v}}, \eta, \bm K, \bm{\sigma}) = \text{EBRPM}(\bm{x_i} = \frac{\sum_{j=1}^p \overline{u}_j \bm{y}_{ji}}{\sum_{j=1}^p \overline{u_j^2}}, \bm S_i = \frac{1}{\sum_{j=1}^p \overline{u_j^2}} \bm{\Sigma_i})
\end{equation}
Note that each $\bm y_{ji}$ is of potentially different dimension, which fits in the framework of the EBRPM model by design. For estimation of the multiple rank model, steps $5,6$ of algorithm \ref{alg:rank_R} and step $3$ of algorithm \ref{alg:greedy} are modified accordingly to accommodate the timepoint structure.

\section{Proofs} \label{sec:proof}
\subsection{Theorem \ref{thrm EB VB equiv}}

We prove Theorem \ref{thrm EB VB equiv}, where we stated the equivalence between step 4, 5 of algorithm \ref{alg:rank_1} and the EBRPM problem. The proof is similar to Proposition 5 in \cite{EBMF}, which can be viewed as a univariate case of our model. We first introduce the following lemma as an intermediate step. 
\begin{supplemma} \label{lemma eb}
For $i=1,\dots,n$, let $\bm{x}_i \overset{iid}{\sim} mvN_{d_i}(\bm{\theta}_i, \bm S_i)$, $\bm{\theta}_i \overset{iid}{\sim} \mathcal{GP}(0,\bm K)$, and $\mathbb{S} = \{\bm S_i:i=1,\dots,n\}$ be known. Recall the EBRPM first estimates the prior variance function as $\hat{\bm K} = \underset{\bm K}{\arg \max} \: \log p(\bm{x} | \bm K)$, then uses this estimate to calculate the posterior $p(\bm{\theta}|\bm{x},\hat{\bm K})$. The solution to the EBRPM problem $(p(\bm{\theta}|\bm{x},\hat{\bm K}), \hat{\bm K})$ is equivalent to the solution of the following objective: 
\begin{equation}
    \underset{q,\bm K}{\arg\max} \: F_0(\bm K,q(\bm{\theta}))
\end{equation}
where
\begin{equation}
    \begin{aligned}
        F_0(\bm K,q(\bm{\theta})) &= E_q[\log{\frac{p(\bm{x},\bm{\theta}|K,\mathbb{S})}{q(\bm{\theta})}}] \\
        &\propto E_{q}[-1/2 \sum_{i=1}^n \bm{\theta}_i^T A_i \bm{\theta}_i - 2B_i^T \bm{\theta}_i] + E_{q} [\log \frac{g(\bm{\theta}|K)}{q(\bm{\theta})}]
    \end{aligned}
\end{equation}

$$\text{where } \bm A_i = \bm S_i^{-1}, \: \bm B_i = S_i^{-1} \bm{x}_i$$
\end{supplemma}
\begin{proof}
    To demonstrate equivalence, we need to show that $\hat{q}(\bm{\theta)} = p(\bm{\theta}|\bm{x},\hat{K})$ and $\underset{K}{\arg \max} \: F_0 = \hat{K}_{EB}$. Note that 
    \begin{equation}
        \begin{aligned}
            \log p(\bm{x}|K) & = \int q(\bm{\theta}) \log \frac{p(\bm{x},\bm{\theta}|K,\bm{S})}{q(\bm{\theta})}d\bm{\theta} + \int q(\bm{\theta}) \log \frac{q(\bm{\theta})}{p(\bm{\theta}|\bm{x},K)}d\bm{\theta} \\
            & = F_0(K,q(\bm{\theta})) + KL(q_\theta||p_{\theta|x})
        \end{aligned}
    \end{equation}
    Since $KL(q_\theta||p_{\theta|x}) \geq 0$ with equality iff $q_\theta = p_{\theta|x}$, we see that
    \begin{equation}
        \begin{aligned}
            \max_{q} F_0(q(\bm{\theta}), K) = \log p(\bm{x}|K)
        \end{aligned}
    \end{equation}
    when $q(\bm{\theta}) = p(\bm{\theta}|\bm{x},K)$, so then 
    \begin{equation}
        \begin{aligned}
            \underset{K}{\arg\max} \{\max_{q} F_0 (q(\bm{\theta}), K)\} = \underset{K}{\arg\max} \log p(\bm{x}|K) = \hat{K}_{EB}
        \end{aligned}
    \end{equation}
    Lastly we show that $F_0$ has the expanded form, which can be done by straightforwardly factorizing the joint distribution:
    
    \begin{equation}
        \begin{aligned}
            F_0(K,q(\bm{\theta})) &= E_q[ \log \frac{p(\bm{x},\bm{\theta}|K,\bm{S})}{q(\bm{\theta})} ] \\
            &= E_q[\log p(\bm{x}| \bm{\theta}, K,\bm{S})] + E_q[\log \frac{g(\bm{\theta}|K)}{q(\bm{\theta})}] \\
            &= E_q[ \sum_{i=1}^n \log N_{d_i}(\bm{x}_i|\bm{\theta}_i,S_i) ] + E_q[\log \frac{g(\bm{\theta}|K)}{q(\bm{\theta})}] \\
            &\propto E_{q}[-1/2 \sum_{i=1}^n \bm{\theta}_i^T A_i \bm{\theta}_i - 2B_i^T \bm{\theta}_i] + E_{q} [\log \frac{g(\bm{\theta}|K)}{q(\bm{\theta})}]
        \end{aligned}
    \end{equation}
\end{proof}
We now prove Thoerem \ref{thrm EB VB equiv}.
\begin{proof}
    In step $4$ of Algorithm \ref{alg:rank_1}, we treat all quantities other than $q_{\bm{u}}, \eta$ as fixed. The objective function is then
    \begin{equation}
        \begin{aligned}
            F(q, \eta, K, \bm{\sigma}) &\coloneq E_{q}[\log \frac{p(Y,u,v; \eta, K,\bm{\sigma})}{q(u,v)}] \\
            &\propto E_{q_{\bm{u}}}[\sum_{t=t_1}^{t_D} \log p(Y_t|\bm{u},\bm{v}_t,\eta,K)] + E_{q_{\bm{u}}}[\log \frac{g(\bm{u}|\eta)}{q(\bm{u})}] \\
            &\propto E_{q_{\bm{u}}}[-1/2 \sum_{j=1}^p \bm{u}_j^T A_j \bm{u}_j - 2B_j^T \bm{u}_j] + E_{q_{\bm{u}}} [\log \frac{g(\bm{u}|\eta)}{q(\bm{u})}]
        \end{aligned}
    \end{equation}
    $$\text{Where } A_j = \sum_{t=t_1}^{t_D} \sum_{i=1}^n \overline{v^2}_{ti} / \sigma_t^2, \: B_j = \sum_{t=t_1}^{t_D} \sum_{i=1}^n Y_{tji} \overline{v}_{ti}/\sigma_t^2$$
    Then equation \ref{alg:1.4 EB} follows from Lemma \ref{lemma eb}.
    Similarly, with respect to $q_{\bm{v}}, K$ for step $5$, the objective function becomes
    \begin{equation}
        \begin{aligned}
            F(q, \eta, K, \bm{\sigma}) \propto E_{q_{\bm{v}}}[-1/2 \sum_{i=1}^n \bm{v}_i^T A_i \bm{v}_i - 2B_i^T \bm{v}_i] + E_{q_{\bm{v}}} [\log \frac{g(\bm{v}|K)}{q(\bm{v})}]
        \end{aligned}
    \end{equation}
    $$\text{Where } A_i = (\sum_{j=1}^p \overline{u_j^2}) \bm{\Sigma}^{-1}, \: B_i = \bm{\Sigma}^{-1} (\sum_{j=1}^p \overline{u}_{j} \bm{y}_{ji}), \: \bm{\Sigma} = \text{diag}(\sigma_{t_1}^2,\dots,\sigma_{t_D}^2)$$
    and equation \ref{alg:1.5 EB} follows from Lemma \ref{lemma eb}.
\end{proof}

\subsection{Theorem \ref{thrm: imputation}}

\begin{proof}
    From algorithm \ref{alg:rank_R}, it can be seen that the M steps of algorithm \ref{alg:impute} each partially maximize the ELBO over $\bm{\sigma}, q(U),q(\mathbb V)$, holding all other quantities fixed. So it leaves to show that the E step maximizes the ELBO over the missing entries $Y_{\mathcal{M}}$. The ELBO can be decomposed into 
    \begin{equation} \label{equn: full ELBO decomp}
        \begin{aligned}
            F_{ELBO} &= E_{q}[\sum_{t=t_1}^{t_D} \log p(Y_t|U,V_t,\bm{\eta},\mathbb{K})] + E_{q_{\bm{u}}}[\log \frac{g(U|\bm{\eta})}{q(U)}] + E_{q_{\bm{v}}}[\log \frac{g(\mathbb{V}|\mathbb{K})}{q(\mathbb{V})}]
        \end{aligned}
    \end{equation}
    holding all other quantities fixed, $F_{ELBO}$ with respect to $Y_{\mathcal{M}}$ is proportional to 
    \begin{equation}
        \begin{aligned}
            F_{ELBO}(Y_{\mathcal{M}}) &\propto E_{q}[\sum_{(t,j,i)\in \mathcal{M}} \log p(Y_{tji}|\bm{u}_j,\bm{v}_{ti},\bm{\eta},\mathbb{K})] \\
            &= \sum_{(t,j,i)\in \mathcal{M}} -\frac{1}{2\sigma_t^2} (Y_{tji} - \sum_{r=1}^R E_q[U_{jr} V_{tir}])^2 + b \leq b \\
            & \text{where $b$ is a constant}
        \end{aligned}
    \end{equation}
    Thus it can be seen that $F_{ELBO}$ is maximized at $Y_{tji} = \sum_{r=1}^R E_q[U_{jr} V_{tir}] = \hat{S}_{tji}$, for all $(t,j,i) \in \mathcal{M}$.
\end{proof}

\section{Solving the EBRPM problem}
\label{app:ebrpm}

Here, we provide the details for implementing the two step solution to solve the EBRPM problem described in Section \ref{sec:EBRPM}. From the model described in equation \ref{equation: EBRPM_1}, \ref{equation: EBRPM_2}, the first step is to estimate $K$ by maximizing the marginal likelihood. We restrict to the case where $K$ is parametrized by some parameters $\bm \alpha$, and estimate $\alpha$ instead, which is the case that all covariance functions discussed fall into. Below, we calculate the analytic form of the gradients and several well-documented gradient-based optimization methods can be used. For convenience, let $K_i = S_i + K$, the marginal log likelihood is given by:

\begin{equation}
    l(\bm \alpha) = -\frac{1}{2} \sum_{i=1}^n \{ d_i \log 2\pi + \sum_{i=1}^n \log |K_i| + \bm x_i^T K_i^{-1} \bm x_i \}
\end{equation}

It can be shown that the matrix derivative of inverse matrices and log determinant of  positive definite matrices are given by
\begin{equation}
    \frac{\partial}{\partial \alpha} K_i^{-1} = -K_i^{-1}\frac{\partial K_i}{\partial \alpha} K_i^{-1}
\end{equation}

\begin{equation}
    \frac{\partial}{\partial \alpha} \log|K_i| = \text{tr}(K_i^{-1})\frac{\partial K_i}{\partial \alpha}
\end{equation}
Where $\frac{\partial K_i}{\partial \alpha}$ denotes the elementwise derivative matrix. Thus the gradient is given by

\begin{equation}
    \begin{aligned}
        \frac{\partial l}{\partial \alpha_j} &= \frac{1}{2} \sum_{i=1}^n \bm x_i^T K_i^{-1} \frac{\partial K_i}{\partial \alpha_j} K_i^{-1} \bm x_i - \text{tr}(K_i^{-1} \frac{\partial K_i}{\partial \alpha_j}) \\
        &= \frac{1}{2} \sum_{i=1}^n \text{tr}\left( (\gamma_i \gamma_i^T - K_i^{-1}) \frac{\partial K_i}{\partial \alpha_j} \right)
    \end{aligned}
\end{equation}
Where $\gamma_i = K_i^{-1} \bm x_i$. Once $\hat{\bm \alpha}$ (and $\hat{K}$) is obtained, the posterior distribution for $\bm \theta_i$ is given by

\begin{equation}
    \bm \theta_i | \bm x_i, \hat{K} \sim \mathcal{N}_{d_i}\left( (S_i^{-1}+\hat{K}^{-1})^{-1}S_i^{-1} \bm x_i, (S_i^{-1}+\hat{K}^{-1})^{-1} \right)
\end{equation}

\section{ELBO Calculation}
\label{app:elbo}

In the CAVI algorithm described in section \ref{sec:EBRPM}, each update step to the parameters does not explicitly compute the objective function itself. Being able to calculate the objective function $F_{ELBO}$ could be beneficial for monitoring the convergence of the optimization algorithm, especially since $F_{ELBO}$ can be highly nonconcave. Here, we provide the procedure for computing $F_{ELBO}$ using the same decomposition as in equation \eqref{equn: full ELBO decomp}. The first term $\sum_{t=t_1}^{t_D} E[\log p(Y_t|U,V_t,\bm \eta, \mathbb{K})]$ is straightforward to compute, given the factor distributions $q(U,\mathbb{V})$. The other two terms can be identified as $\sum_{r=1}^R -KL(q(U_r) \| g(U_r))$ and $\sum_{r=1}^R -KL(q(\mathbb{V}_r) \| g(\mathbb{V}_r))$, each $r$'th term from either sums can be found as follows from the EBRPM subproblem:

\begin{equation}
    \begin{aligned}
        \log p(\bm x | \hat{K}) &= F_{0}(\hat{K},\hat{q}(\bm \theta)) = E_q[\log{\frac{p(\bm{x},\bm{\theta}|\hat{K},\bm{S})}{\hat{q}(\bm{\theta})}}] \\
        &= E_q[\log p(\bm x|\bm \theta)] + E_q[\log \frac{\hat{g}(\bm \theta)}{\hat{q}(\bm \theta)}] \\
        E_q[\log \frac{\hat{g}(\bm \theta)}{\hat{q}(\bm \theta)}] &= l(\hat{g}) + 1/2 \sum_{i=1}^n \Bigl\{ d_i\log(2\pi) + \log|S_i| \\ 
        &+ \left(\bm x_i^T S_i^{-1} \bm x_i - 2 \bm x_i^T S_i^{-1} E_q[\bm \theta_i] + \text{tr}(S_i^{-1}Var_q(\bm \theta_i)) + E_q[\bm \theta_i]^T S_i^{-1} E_q[\bm \theta_i] \right) \Bigl\}
    \end{aligned}
\end{equation}

This calculation is straightforward to implement, as the only additional quantity required is the optimized log likelihood value from the EBRPM problem, which is usually not difficult to obtain.

\section{Simulations: Data Generation}
\label{app:sims}

In the simulation section, we set the sample size $n=120$, variable dimension $p=100$, and number of timepoints to be $D=5$ complete data are generated according to the prior described in \eqref{equn: prior}: For $r = 1, \dots, R$, draw the factor matrices $\bm u_r \sim N(0, \eta^2 I)$, $\tilde{\bm{v}}_{ri} \sim N(0,\bm K)$, compute the underlying structure $\bm S_t = \bm U \bm V^T_t$ and $\bm X_t = \bm S_t + \bm E_t$ where entries of $\bm E_t$ are standard normal draws. For the covariance structures \eqref{cov.ec}, \eqref{cov.rbf}, we set $\sigma_v = 0.2,0.5,1$ for the three SNR we consider and fix $\eta$ at $1$. {\chLB This gives the low, medium, high SNR settings the SNR of $0.12, 0.75, 3$ respectively, as the total SNR is the sum of each single rank component's variance.} In low and high correlation levels, we set $\rho=0.2, 0.8$ for the $EC$ structure and $\alpha=1,3$ for the $RBF$ structure respectively. The free covariance structure \eqref{free:covariance} is generated as $\bm A^T \bm A/{\chLB d}$, where $\bm A$ has standard normal entries. The block covariance is defined as a block diagonal covariance matrix $diag(\bm \Sigma_1, \bm \Sigma_2)$ where $\bm \Sigma_1, \bm \Sigma_2$ are 2-by-2 and 3-by-3 $EC$ covariances respectively.

\section{{\chLB Additional Simulations}}

\subsection{Irregular timepoints}

\subsubsection{Settings}

In this experiment we investigate our model's performance under irregular timepoints. The data dimensions are set to $n=120$, $p=100$, $R=3$, consistent with Section~\ref{Sec: Simulation}. The loadings and scores are generated as described in section~\ref{app:sims}. For subject $i$, each score vector is drawn from the \textit{rbf} covariance kernel. Parameters of the kernel are set to scale $\eta \in \{1, 0.5, 0.2\}$ and length-scale $\alpha \in \{3, 1\}$ for all of the $R$ components. The noise variance is set to $\sigma=1$. Each setting is repeated $100$ times.

\begin{figure}[ht]
\centering
\includegraphics[width=\textwidth]{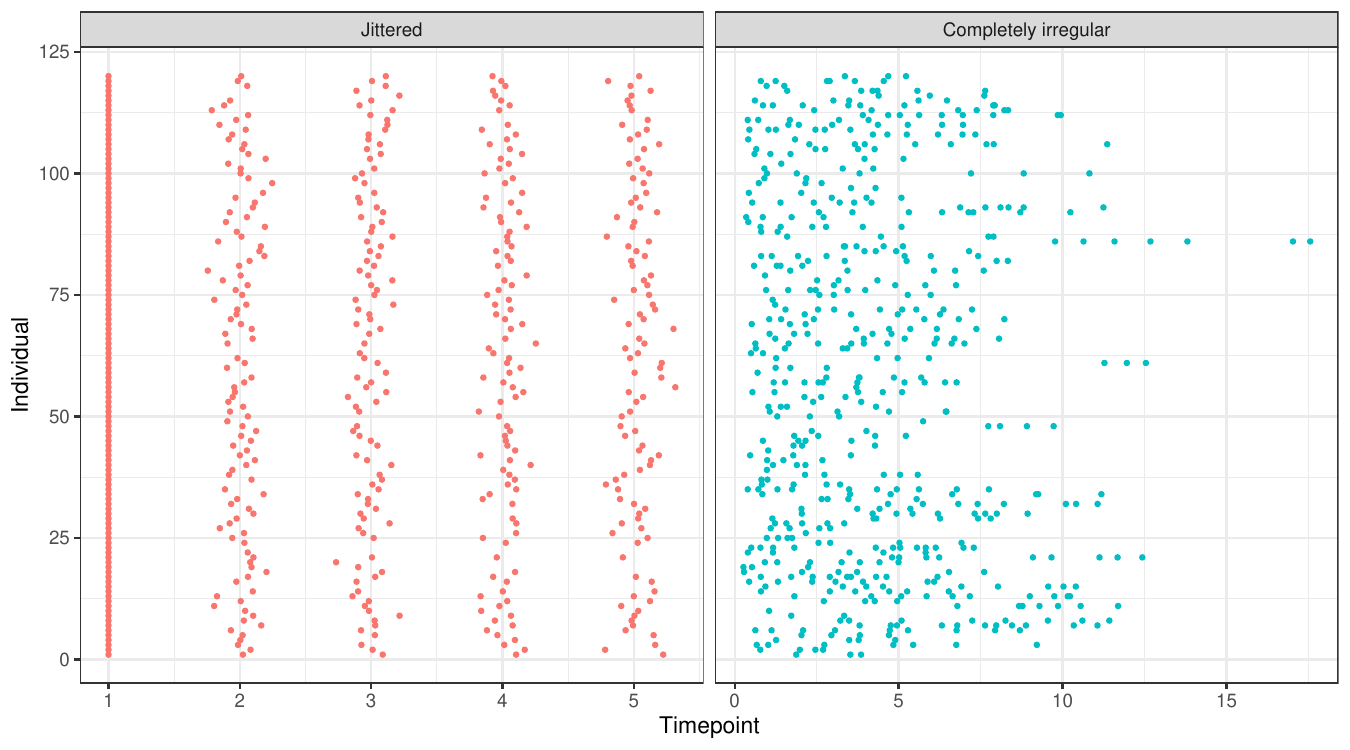}
\caption{Visit times for 120 subjects in one draw from each design.}
\label{fig:irregular_times}
\end{figure}

We used two designs for the visit times (Figure~\ref{fig:irregular_times}).
\begin{itemize}
\item \textbf{Jittered.} Each subject has five planned fixed timepoints at times $1,\dots,5$. Each visit is then offset by a small amount of noise $N(0, 0.1^2)$. We then randomly remove $7\%$ of the visits. After each visit, the subject drops out with probability $0.03$.
\item \textbf{Completely irregular.} The number of timepoints is generated from a negative binomial distribution with size $2.8$ and mean $4$ then added $1$, upper bounded at 15 timepoints. The time of the first visit is randomly generated using a $\mathrm{LogNormal}(0.4, 0.83)$ distribution, followed by the time between visits, randomly drawn from a $\mathrm{Gamma}(2,2)$ distribution.
\end{itemize}

For each dataset, 10\% of all visits are held out, one visit for each selected subject. Each subject is guaranteed to have at least one timepoint, so we only randomly sample from subjects with two or more timepoints. We fitted two types of model:
\begin{itemize}
\item \textbf{Irregular time model.} The irregular model uses the true visit times of each subject. We used the $EC$ and $RBF$ priors.
\item \textbf{Regular time model with binned times.} Observed timepoints are rounded to the nearest whole time unit. If multiple visits are in one time bin, we take the average. We then fitted the aligned model with the $EC$, $Free$ and $RBF$ priors.
\end{itemize}

Figure~\ref{fig:irregular_times} provides a visualization of the generated timepoints from a single dataset. To assess performance, we report the relative squared error $\|\hat{S} - S\|_F^2 / \|S\|_F^2$ at the observed visits and at the held-out visits. We also report the selected rank and the fit time.

\subsubsection{Results}

Table~\ref{tab:irr_obs} and Table~\ref{tab:irr_held} provides a summary of the RSE performance on observed visits and heldout visits. In the jittered time design, both the irregular time model and the regular time model fitted with binned timepoints perform comparably across all SNR and correlation strength settings. However, in the irregular time setting, fitting the regular time model fails severely, especially when SNR is low. The irregular time models selected the true rank in all simulation replicates across all settings. On the other hand, while the binned model selected the true rank in all jittered time settings, Table~\ref{tab:irr_rank} shows that it was unable to accurately estimate the rank in all completely irregular time settings with $\eta = 0.2$. Table~\ref{tab:irr_rank} gives the settings where the binned models failed. These results demonstrate that correctly accounting for the irregular timepoints is essential and that fitting the regular time model with approximated timepoints results in poor model fit and incorrect model structure.

\begin{table}[!ht]
\centering
\small
\caption{Mean RSE at the observed visits.}
\label{tab:irr_obs}
\begin{tabular}{lccccccc}
\hline
Design & $\eta$ & $\alpha$ & EC irregular & RBF irregular & EC binned & Free binned & RBF binned \\
\hline
Jittered & 1 & 3 & 0.012 & 0.009 & 0.012 & 0.010 & 0.009 \\
 & 1 & 1 & 0.012 & 0.012 & 0.012 & 0.012 & 0.012 \\
 & 0.5 & 3 & 0.044 & 0.032 & 0.045 & 0.034 & 0.033 \\
 & 0.5 & 1 & 0.047 & 0.045 & 0.047 & 0.048 & 0.046 \\
 & 0.2 & 3 & 0.200 & 0.159 & 0.212 & 0.166 & 0.161 \\
 & 0.2 & 1 & 0.251 & 0.232 & 0.255 & 0.248 & 0.238 \\
\hline
Irregular & 1 & 3 & 0.012 & 0.008 & 0.024 & 0.014 & 0.016 \\
 & 1 & 1 & 0.012 & 0.011 & 0.039 & 0.032 & 0.040 \\
 & 0.5 & 3 & 0.044 & 0.028 & 0.143 & 0.046 & 0.053 \\
 & 0.5 & 1 & 0.046 & 0.040 & 0.110 & 0.076 & 0.105 \\
 & 0.2 & 3 & 0.212 & 0.147 & 0.838 & 0.276 & 0.675 \\
 & 0.2 & 1 & 0.245 & 0.203 & 0.995 & 0.460 & 0.925 \\
\hline
\end{tabular}
\vspace{1em}

\caption{RSE at the held-out visits.}
\label{tab:irr_held}
\begin{tabular}{lccccccc}
\hline
Design & $\eta$ & $\alpha$ & EC irregular & RBF irregular & EC binned & Free binned & RBF binned \\
\hline
Jittered & 1 & 3 & 0.276 & 0.040 & 0.280 & 0.044 & 0.043 \\
 & 1 & 1 & 0.844 & 0.432 & 0.842 & 0.465 & 0.440 \\
 & 0.5 & 3 & 0.297 & 0.082 & 0.297 & 0.088 & 0.084 \\
 & 0.5 & 1 & 0.841 & 0.474 & 0.840 & 0.530 & 0.486 \\
 & 0.2 & 3 & 0.365 & 0.238 & 0.365 & 0.252 & 0.242 \\
 & 0.2 & 1 & 0.861 & 0.616 & 0.861 & 0.669 & 0.623 \\
\hline
Irregular & 1 & 3 & 0.338 & 0.065 & 0.255 & 0.136 & 0.083 \\
 & 1 & 1 & 0.812 & 0.359 & 0.604 & 0.565 & 0.457 \\
 & 0.5 & 3 & 0.332 & 0.095 & 0.274 & 0.158 & 0.132 \\
 & 0.5 & 1 & 0.784 & 0.380 & 0.630 & 0.625 & 0.488 \\
 & 0.2 & 3 & 0.410 & 0.238 & 0.856 & 0.371 & 0.702 \\
 & 0.2 & 1 & 0.826 & 0.524 & 0.998 & 0.786 & 0.959 \\
\hline
\end{tabular}
\vspace{1em}

\caption{Number of replicates (out of 100) with the correct rank for the binned models, completely irregular design, $\eta = 0.2$. The range of the selected rank is included in parentheses.}
\label{tab:irr_rank}
\begin{tabular}{lccc}
\hline
$\alpha$ & EC binned & Free binned & RBF binned \\
\hline
3 & 0 (0--1) & 76 (2--3) & 0 (0--2) \\
1 & 0 (0--1) & 39 (1--3) & 0 (0--1) \\
\hline
\end{tabular}
\end{table}

\subsection{Misspecified Settings}
\label{sec:misspec}

In this section, we investigate the robustness of our model to several ways in which data generation can misalign with the model. Unless stated otherwise, we use $n=120$, $p=100$, $5$ timepoints, $\eta \in \{1, 0.5, 0.2\}$, $\sigma=1$, $10\%$ entrywise or columnwise missingness and $100$ replicates per setting, as in the main simulations.

\subsubsection{Tensor data structure}

\begin{figure}[ht]
\centering
\includegraphics[width=\textwidth]{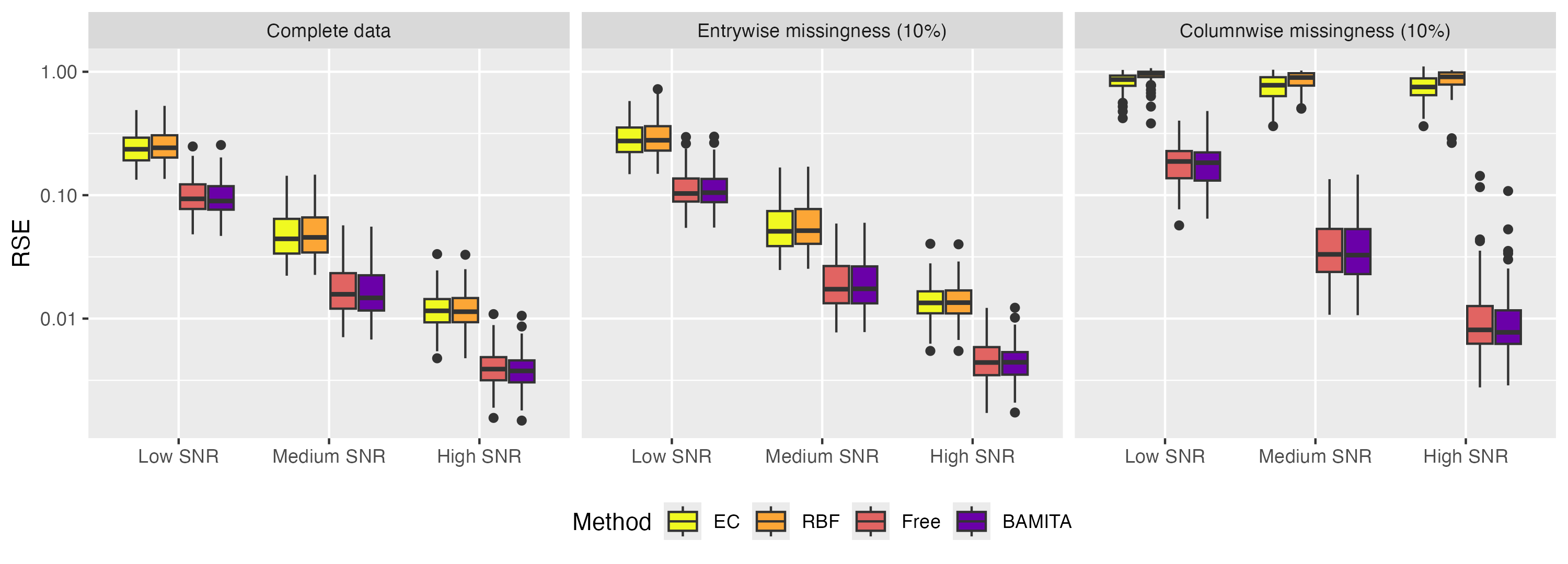}
\caption{RSE for data generated from the CP tensor model.}
\label{fig:misspec_cp}
\end{figure}

Let $\circ$ denote the outer product. Data are generated from a rank-$3$ CP tensor model,
\begin{equation}
    \mathcal{X} = \sum_{r=1}^{3} \mathbf{u}_r \circ \mathbf{v}_r \circ \mathbf{w}_r + \mathcal{E},
\end{equation}
where $\mathbf{u}_r \in \mathbb{R}^{n}$ and $\mathbf{v}_r \in \mathbb{R}^{p}$ have $N(0,1)$ entries, $\mathbf{w}_r \in \mathbb{R}^{5}$ has $N(0,\eta^2)$ entries to control the SNR, and the random errors terms $\mathcal{E}$ are iid distributed $N(0,1)$. Figure~\ref{fig:misspec_cp} shows the results under completely observed data, $10\%$ random entrywise missingness, and $10\%$ random columnwise missingness. In all three settings, we observe that while the $EC$ and $RBF$ models perform worse than the correctly specified BAMITA model, the $Free$ covariance model performs comparably across all SNR and observed/missing data settings. Indeed, the tensor structure is very similar to the $Free$ covariance model, as each frontal slice of the tensor can be viewed as a matrix observation at a single timepoint. Under this angle, the flexible $Free$ covariance structure model is able to capture covariance that is not structured smoothly over time, demonstrating robustness to such misspecifications.

\subsubsection{Mixture of temporal \& static factors}

\begin{figure}[ht]
\centering
\includegraphics[width=\textwidth]{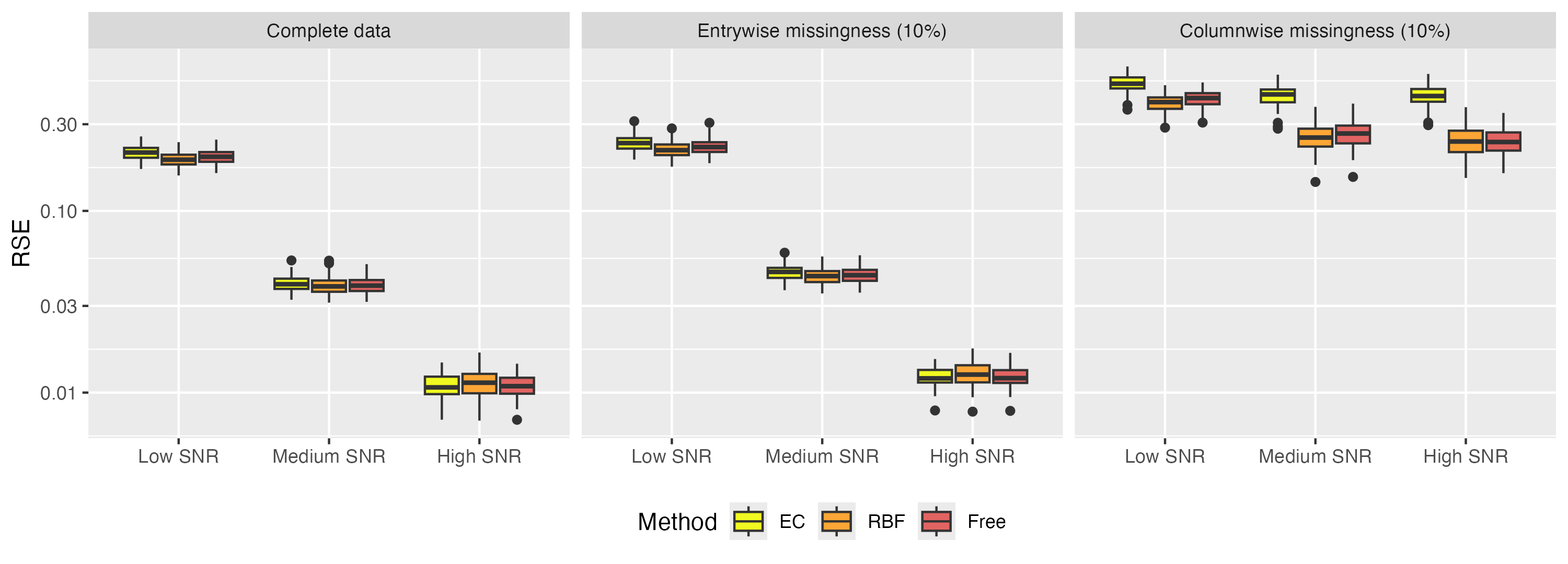}
\caption{RSE for data with two static and two temporal factors.}
\label{fig:misspec_mix}
\end{figure}

Data are generated from our model with rank $R=4$: two static factors, whose scores are the same at every timepoint for each subject, and two temporal factors, whose scores follow the RBF covariance with $\alpha=1$. Each factor has score variance $3\eta^2/4$, so the total signal matches the rank-$3$ settings. From Figure~\ref{fig:misspec_mix}, we observe that our model performs similar to the correctly specified setting, and captures the true signal of the model.

\subsubsection{Non-Gaussian loadings}

\begin{figure}[ht]
\centering
\includegraphics[width=\textwidth]{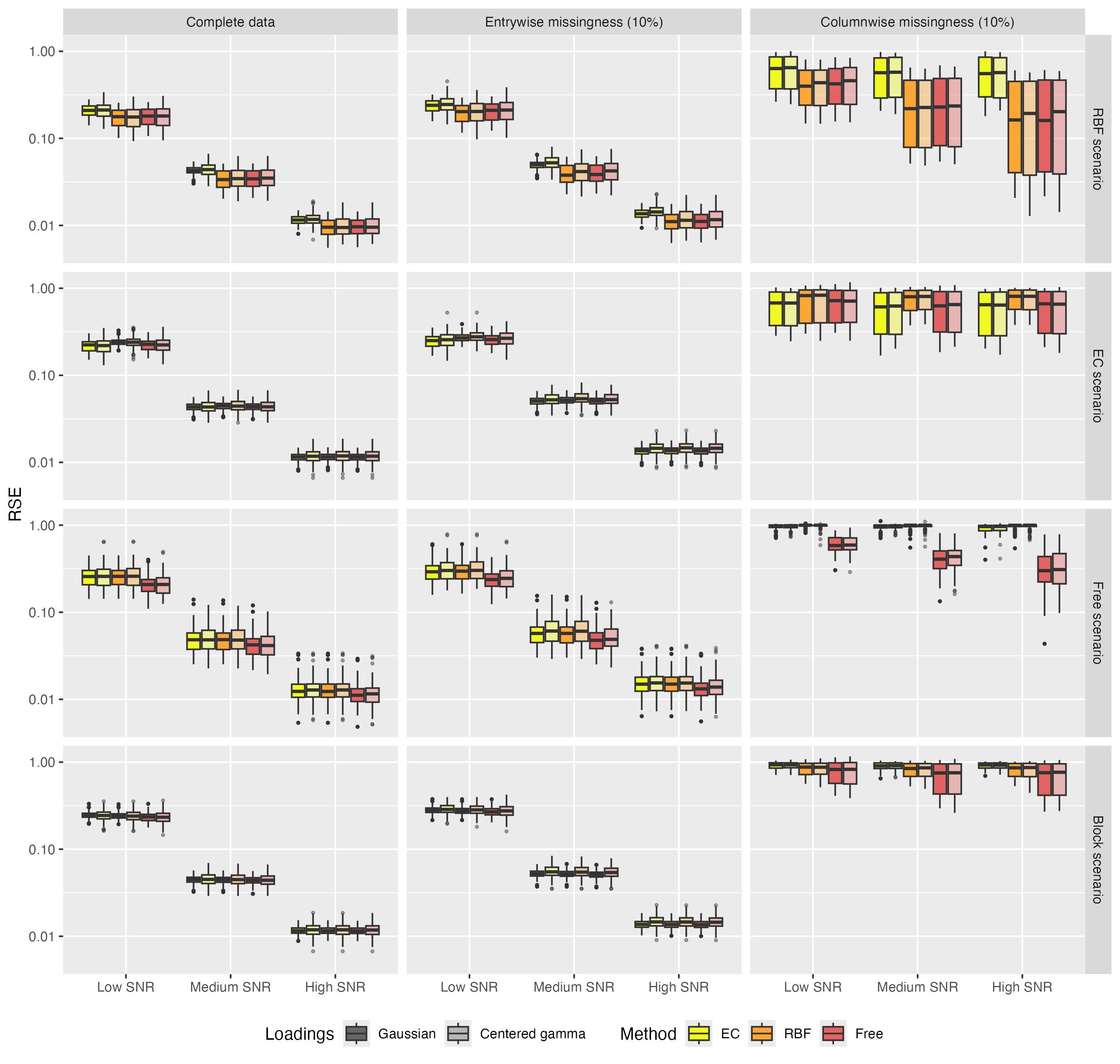}
\caption{RSE with centered gamma loadings (faded) and Gaussian loadings (solid). Rows give the true covariance of the scores.}
\label{fig:misspec_gamma}
\end{figure}

Data are generated as in the main simulations from Section~\ref{Sec: Simulation}, with the loading-generating distribution replaced by the centered $\mathrm{Gamma}(1,1)$ distribution. Each setting is compared with the same setting under loadings generated from a standard normal distribution. Figure~\ref{fig:misspec_gamma} shows that the models under non-Gaussian loadings perform similarly to the correctly specified Gaussian loading setting, and thus misspecification of the loading distribution does not result in poor model fit of the EVBLM model.

\subsection{Scalability}
\label{sec:scalability}

In this experiment, we investigate the scalability of our model. We benchmark the runtime to fit our model by varying each of the number of subjects, variables, timepoints, and missingness. Data are generated from our model with the \textit{rbf} covariance with parameters $\eta=0.5$, $\alpha=3$, and $\sigma^2=1$. We establish a reference setting of $p=1000$ variables, $n=200$ subjects, $T=5$ timepoints and rank $3$. We scale one of the dimensions each at a time: from $100$ to $20000$ variables; $50$ to $5000$ subjects; $3$ to $50$ timepoints; rank $1$ to $20$; and the total entries at a timepoint $(p\times n)$ from $250 \times 50$ to $(10000 \times 2000)$, which gives $6.25 \times 10^4$ to $10^8$ entries; and $10\%$, $30\%$ and $50\%$ of the entries or of the visits missing. Each setting is repeated 100 times. The model is fitted with the \textit{ec}, \textit{free} and \textit{rbf} covariances, with automatic rank selection. All timings were obtained on a single node with two 64-core AMD EPYC 7763 processors and 512 GB of RAM, using R 4.3.0 with single-threaded OpenBLAS, with eight fits running concurrently. 

\begin{figure}[ht]
\centering
\includegraphics[width=\textwidth]{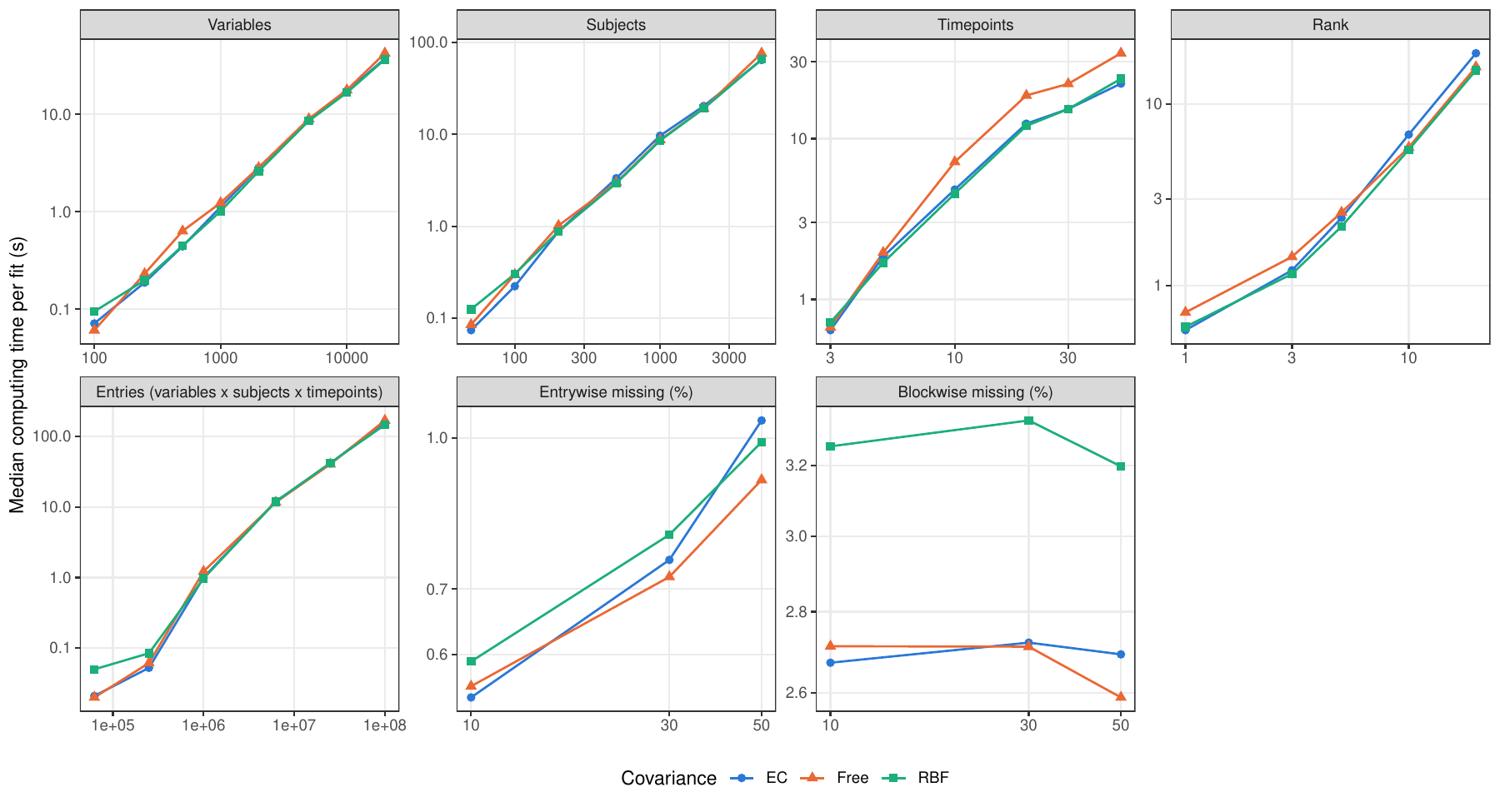}
\caption{Median computing time per fit (100 replicates) as each dimension increases, with the other dimensions fixed at 1000 variables, 200 subjects, 5 timepoints and rank 3. Both axes are on a log scale.}
\label{fig:scaling_time}
\end{figure}

Figure~\ref{fig:scaling_time} shows the median runtime per model fit. The log computing time is linear in log growth of each of $p, n,$ and $D$. The consistent behavior is expected as these parameters all scale the total number of matrix entries linearly. This is also observed in the first panel on the second row of Figure~\ref{fig:scaling_time}, where we scale the parameters by the product of $p,n,D$. All three covariance models perform comparably when scaling $p$ and $n$. The \textit{free} covariance model takes longer to fit with more timepoints, as there are $D(D+1)/2$ parameters to estimate, whereas the parameters of \textit{EC} and \textit{rbf} are constant with respect to timepoints. In the missingness scenarios, the computing time depends on the mode of missingness. Entrywise missingness scales the log computing time linearly similar to the previous settings, because more missing entries require more iterations of the imputation step before the ELBO converges. On the other hand, computing time stays relatively stable in the blockwise missingness scenario, and even faster when $50\%$ of the visits are missing.

\section{IFN\texorpdfstring{$\beta$}{β} Data Application}

\subsection{{\chLB Preprocessing}}
\label{sec:ifnb_preproc}

As described in the original study \cite{IFNB}, expression of pre-selected genes was measured in PBMC RNA by one-step kinetic RT-PCR (ABI 7900), with triplicate reactions on 384-well plates that included positive and negative controls and a calibration curve. Absolute expression values were obtained by interpolation from a calibration curve spanning five orders of magnitude (built from an in vitro transcribed GAPDH clone), and the triplicate means were divided by the value of a positive-control RNA (thymus) on the same plate to correct for plate-to-plate variability. Housekeeping-gene normalization was deliberately not used; instead, equal RNA input was ensured through fluorescence-based RNA quantitation, dilution of all samples to a common concentration, and automated liquid handling. The original authors did not apply a formal batch correction beyond this plate-level normalization. We used the publicly released, normalized expression values (Dataset S1), which were not log-scaled, applied a natural log transformation, and then centered each gene to have mean zero across its observed values. No further processing was performed.

\subsection{{\chLB Number of components}}

The greedy algorithm adds one component at a time and stops when a new component does not increase the ELBO; backfitting then updates all components jointly. With the EC covariance this selects 13 components on the full data. Table~\ref{tab:ifnb_boot} shows the number selected across bootstrap resamples (Section~\ref{sec:ifnb_boot}).

\subsection{ {\chLB Responder Difference Significance Test}}
\label{sec:ifnb_lmm}

Let $S_{ij}$ be the posterior mean score of component 7 for subject $i$ at visit $j$, with $t_j \in \{0, 0.25, 0.5, 0.75, 1, 1.5, 2\}$ the visit time in years from baseline, and $R_i$ the indicator that subject $i$ is a good responder. We fit the linear mixed model
\begin{equation}
S_{ij} = \beta_0 + \beta_1 R_i + \beta_2 t_j + \beta_3 R_i t_j + e_i + \varepsilon_{ij}, \qquad e_i \sim N(0, \tau^2), \quad \varepsilon_{ij} \sim N(0, \sigma^2),
\label{eq:ifnb_lmm}
\end{equation}
where $e_i$ is a random subject intercept. Here $\beta_1$ is the difference between good and poor responders at baseline, and $\beta_3$ is the change in that difference per year; if $\beta_3 = 0$, the baseline difference holds at every visit. The model uses all $53 \times 7$ estimated scores, including those at missing visits, and is fit by REML with the \texttt{nlme} package.

The estimates are $\hat\beta_1 = 0.364$ (SE $0.054$, $p = 1.8 \times 10^{-8}$) and $\hat\beta_3 = 0.0006$ per year (SE $0.0041$, $p = 0.89$). Good responders have higher component 7 scores, and we find no evidence that the difference changes over time. These $p$-values treat the estimated scores as observed and do not account for component 7 having been chosen after inspecting the score trajectories; a Bonferroni correction over the 13 components gives $p = 2.3 \times 10^{-7}$ for $\beta_1$.

\subsection{{\chLB Stability across resamples}}
\label{sec:ifnb_boot}

The EVBLM initialization is deterministic, so refitting the full data returns the same components; we therefore assess stability over resamples of subjects. We drew $B = 1000$ bootstrap samples of the 53 subjects with replacement and refit EVBLM with the same settings (EC covariance, greedy selection followed by backfitting). Because components with similar variance explained can change order between fits, the component matched to full-data component 7 is the one whose loading vector has the largest absolute correlation with the full-data loading of component 7. Its scores are multiplied by the sign of that correlation, and model \eqref{eq:ifnb_lmm} is refit to them.

\begin{table}[ht]
\centering
\caption{Bootstrap results for the IFN$\beta$ data ($B = 1000$): number of components selected, and position of the component matched to full-data component 7.}
\label{tab:ifnb_boot}
\begin{tabular}{lrrrrrrrrrrr}
\hline
Number of components & 12 & 13 & 14 & 15 & 16 & 17 & & & & & \\
Count & 10 & 191 & 521 & 253 & 24 & 1 & & & & & \\
\hline
Matched position & 2 & 3 & 4 & 5 & 6 & 7 & 8 & 9 & 10 & 11 & 13 \\
Count & 3 & 36 & 46 & 104 & 173 & 407 & 162 & 40 & 16 & 9 & 4 \\
\hline
\end{tabular}
\end{table}

The median absolute loading correlation of the matched component is $0.84$ (interquartile range $0.72$ to $0.90$), and it exceeds $0.8$ in $61\%$ of resamples. The matched component is at position 5 to 8 in $85\%$ of resamples. Across resamples, $\hat\beta_1$ has median $0.352$ (2.5\% and 97.5\% quantiles $0.122$ and $0.473$) and $p < 0.05$ in $96.8\%$ of resamples; $\hat\beta_3$ has median $0.000$ (quantiles $-0.015$ and $0.019$) and $p < 0.05$ in $10.6\%$.

\end{document}